\documentclass[twoside,11pt]{article}

\usepackage[preprint]{jmlr2e}

\usepackage{subcaption}
\usepackage{booktabs}
\usepackage{url}
\usepackage{multirow}
\usepackage{tikz}
\usetikzlibrary{calc}
\usetikzlibrary{decorations.pathmorphing}
\usepackage{csquotes}
\usepackage{paralist}

\pgfdeclarelayer{background}
\pgfsetlayers{background,main}

\usetikzlibrary{shapes}
\usetikzlibrary{arrows.meta,backgrounds,automata}
\usetikzlibrary{positioning,shapes,matrix,calc}
\tikzstyle{vertex}=[circle, draw, fill=gray!80!white,thick,scale=1.2]
\tikzstyle{edge}=[draw=black, thick,-]

\usepackage{bm}

\definecolor{purple}{RGB}{147,7,204}
\definecolor{green}{RGB}{5,100,18}
\definecolor{blue}{RGB}{10,153,201}
\definecolor{orange}{RGB}{254,128,41}
\definecolor{gray}{RGB}{239,240,241}

\newcommand{\oms}{\{\!\!\{}
\newcommand{\cms}{\}\!\!\}}

\usepackage{thmtools}		
\usepackage{mleftright}
\usepackage{nicefrac}

\usepackage{thm-restate}
\usepackage{mathtools}
\usepackage{fixmath}
\usepackage{siunitx}
\usepackage{color}

\usepackage{pifont}

\usepackage{setspace}
\usepackage[]{microtype}
\usepackage{ellipsis}
\usepackage{xspace}

\usepackage{tcolorbox}

\makeatletter
\def\thmt@refnamewithcomma #1#2#3,#4,#5\@nil{%
	\@xa\def\csname\thmt@envname #1utorefname\endcsname{#3}%
	\ifcsname #2refname\endcsname
	\csname #2refname\expandafter\endcsname\expandafter{\thmt@envname}{#3}{#4}%
	\fi
}
\makeatother
\usepackage[capitalise,noabbrev]{cleveref}   

\newcommand{\new}[1]{\emph{#1}}

\newcommand{\bbR}{\ensuremath{\mathbb{R}}}

\newcommand{\bbZ}{\ensuremath{\mathbb{Z}}}

\newcommand{\RR}{\mathbb{R}}

\newcommand{\NN}{\mathbb{N}}

\ShortHeadings{Combinatorial Optimization and Reasoning with GNNs}{Cappart, Chételat, Khalil, Lodi, Morris, Veli\v{c}kovi\'{c}}
\firstpageno{1}

\jmlrheading{1}{2021}{1-42}{4/21}{4/21}{meila00a}{Quentin Cappart, Didier Chételat, Elias B. Khalil, Andrea Lodi, Christopher Morris, Petar Veli\v{c}kovi\'{c}}

\firstpageno{1}

\newcommand{\an}[1]{}

\begin{document}

\title{Combinatorial Optimization and Reasoning\\ with Graph Neural Networks}

\author{\name Quentin Cappart  \email quentin.cappart@polymtl.ca \\
       \addr Department of Computer Engineering and Software Engineering\\
       Polytechnique Montr{\'e}al\\
       Montr{\'e}al, Canada
       \AND
       \name Didier Chételat  \email didier.chetelat@polymtl.ca \\
       \addr CERC in Data Science for Real-Time Decision-Making\\
       Polytechnique Montr{\'e}al\\
       Montr{\'e}al, Canada
       \AND
       \name Elias B. Khalil \email khalil@mie.utoronto.ca\\
       \addr Department of Mechanical \& Industrial Engineering,\\
       University of Toronto\\
       Toronto, Canada
       \AND
       \name Andrea Lodi  \email andrea.lodi@cornell.edu \\
       \addr Jacobs Technion-Cornell Institute\\
       Cornell Tech and Technion - IIT\\
       New York, USA
        \AND
       \name  Christopher Morris  \email morris@cs.rwth-aachen.de \\
       \addr Department of Computer Science\\
       RWTH Aachen University\\
       Aachen, Germany
      \AND
       \name Petar Veli\v{c}kovi\'{c} \email petarv@deepmind.com\\
       \addr DeepMind\\
       London, UK
     }

\editor{TBD}

\maketitle

\begin{abstract}%
Combinatorial optimization is a well-established area in operations research and computer science. Until recently, its methods have focused on solving problem instances in isolation, ignoring that they often stem from related data distributions in practice. However, recent years have seen a surge of interest in using machine learning, especially graph neural networks (GNNs), as a key building block for combinatorial tasks, either directly as solvers or by enhancing exact solvers. The inductive bias of GNNs effectively encodes combinatorial and relational input due to their invariance to permutations and awareness of input sparsity. This paper presents a conceptual review of recent key advancements in this emerging field, aiming at optimization and machine learning researchers.
\end{abstract}


\section{Introduction}

Combinatorial optimization (CO) has developed into an interdisciplinary field spanning optimization, operations research, discrete mathematics, and computer science, with many critical real-world applications such as vehicle routing or scheduling; see~\citep{Kor+2012} for a general overview. Intuitively, CO deals with problems that involve optimizing a cost (or objective) function by selecting a subset from a finite set, with the latter encoding constraints on the solution space. Although CO problems are generally hard from a complexity theory standpoint due to their discrete, non-convex nature~\citep{karp1972reducibility},
many of them are routinely solved in practice. Historically, the optimization and theoretical computer science communities have been focusing on finding optimal~\citep{Kor+2012}, heuristic~\citep{boussaid2013survey}, or approximate~\citep{Vaz+2010} solutions for individual problem instances. However, in many practical situations of interest, one often must solve problem instances with specific characteristics or patterns. For example, a trucking company may solve vehicle routing instances for the same city daily, with only slight differences across instances in the travel times due to varying traffic conditions. Hence, data-dependent algorithms or machine learning approaches, which may exploit these patterns, have recently gained traction in the CO field~\citep{Bengio2018,ML4CO}. The promise here is that one can develop faster algorithms for practical cases by exploiting common patterns in the given instances.

Due to the discrete nature of most CO problems and the prevalence of network data in the real world, graphs are a central object of study in the CO field. For example, well-known and relevant problems such as the Traveling Salesperson problem (TSP) and other vehicle routing problems naturally induce a graph structure. In fact, from the 21 \textsf{NP}-complete problems identified by \citet{karp1972reducibility}, ten are decision versions of graph optimization problems.  Most of the other ones, such as the set covering problem, can also be modeled over graphs. Moreover, the interaction between variables and constraints in combinatorial optimization problems naturally induces a bipartite graph, i.e., a variable and constraint share an edge if the variable appears with a non-zero coefficient in the constraint. These graphs commonly exhibit patterns in their structure and features, which machine learning approaches should exploit.

\subsection{What are the Challenges for Machine Learning?}
There are several critical challenges in successfully applying machine learning methods within CO, especially for problems involving graphs. Graphs have no unique representation, i.e., renaming or reordering the nodes does not result in different graphs. Hence, for any machine learning method dealing with graphs, taking into account invariance to permutation is crucial. Combinatorial optimization problem instances are large and usually sparse, especially those arising from the real world. Hence, the employed machine learning method must be scalable and sparsity aware. Simultaneously, the employed method has to be expressive enough to detect and exploit the relevant patterns in the given instance or data distribution. The machine learning method should be capable of handling auxiliary information, such as objective and user-defined constraints.
Most of the current machine learning approaches are within the supervised regime. That is, they require a large amount of training data to optimize the model's parameters. In the context of CO, this means solving many possibly hard problem instances, which might prohibit the application of these approaches in real-world scenarios. Further, the machine learning method has to be able to generalize beyond its training data, e.g., transferring to instances of different sizes.

Overall, there is a trade-off between scalability, expressivity, and generalization, any pair of which might conflict. In summary, the key challenges are:
\begin{enumerate}
	\item Machine learning methods that operate on graph data have to be \emph{invariant} to node \emph{permutations}. They should also exploit the graph's \emph{sparsity}.
	\item Models should \emph{distinguish} critical structural \emph{patterns} in the provided data while still \emph{scaling} to large real-world instances.
	\item \emph{Side information} in the form of high-dimensional vectors attached to nodes and edges, i.e., modeling objectives and additional information, need to be considered.
	\item Models should be \emph{data efficient}. That is, they should ideally work without requiring large amounts of labeled data, and they should be transferable to \emph{out-of-sample} or \emph{out-of-distribution} instances.
\end{enumerate}

\subsection{How Do GNNs Address These Challenges?}
\emph{Graph neural networks} (GNNs)~\citep{Gil+2017,Sca+2009} have recently emerged as machine learning architectures that partially address the challenges above.

The key idea underlying GNNs is to compute a vectorial representation, e.g., a real vector, of each node in the input graph by iteratively aggregating features of neighboring nodes. The GNN is trained in an end-to-end fashion against a loss function, using (stochastic) first-order optimization techniques to adapt to the given data distribution by parameterizing this aggregation step. The promise here is that the learned vector representation encodes crucial graph structures that help solve a CO problem more efficiently. GNNs are invariant and equivariant by design, i.e., they automatically exploit the invariances or symmetries inherent to the given instance or data distribution. Due to their local nature, by aggregating neighborhood information, GNNs naturally exploit sparsity, leading to more scalable models on sparse inputs. Moreover, although scalability is still an issue, they scale linearly with the number of edges and employed parameters, while taking  multi-dimensional node and edge features into account~\citep{Gil+2017}, naturally exploiting cost and objective function information. However, the data-efficiency question is still largely open~\citep{Mor+2021}.

Although GNNs have clear limitations, which we will also explore and outline, they have already proven to be useful in the context of CO. In fact, they have already been applied in various settings, either to directly predict a solution or as an integrated component of an existing solver. We will extensively investigate both of these aspects within our survey.

Perhaps one of the most widely publicized applications of GNNs in CO at the time of writing is the work by \citet{mirhoseini2020chip}, which studies chip placement. The aim is to map the nodes of a \emph{netlist} (the graph describing the desired chip) onto a chip canvas (a bounded 2D space), optimizing the final power, performance, and area. The authors observe this as a combinatorial problem and tackle it using reinforcement learning. Owing to the graph structure of the netlist, at the core of the representation learning pipeline is a GNN, which computes node features in a (permutation-)invariant way. This represents the first chip placement approach that can quickly generalize to previously unseen netlists, generating optimized placements for Google's TPU accelerators \citep{jouppi2017datacenter}. While this approach has received wide coverage in the popular press, we believe that it only scratches the surface of the innovations that can be enabled by a careful synergy of GNNs and CO. We have designed our survey to facilitate future research in this emerging area. 

\subsection{Going Beyond Classical Algorithms}

The previous discussion mainly dealt with the idea of machine learning approaches, especially GNNs, replacing and imitating classical combinatorial algorithms or parts of them, potentially adapting better to the specific data distribution of naturally-occurring problem instances. However, classical algorithms heavily depend on human-made pre-processing or feature engineering by abstracting raw, real-world inputs, e.g., specifying the underlying graph itself. The discrete graph input, forming the basis of most CO problems, is seldomly directly induced by the raw data, requiring costly and error-prone {feature engineering}. This might lead to biases that do not align with the real world and consequently imprecise decisions. Such issues have been known as early as the 1950s in the context of railways network analysis \citep{harris1955fundamentals}, but remained out of the spotlight of theoretical computer science that assumes problems are abstractified, to begin with.

In the long-term, machine learning approaches can further enhance the CO pipeline, from raw input processing to aiding in solving abstracted CO problems in an end-to-end fashion. Several viable approaches in this direction have been proposed recently, and we will survey them in detail, along with motivating examples, in Section \ref{subsub:rni}.

\subsection{Present Work} In this paper, we give an overview of recent advances in using GNNs in the context of CO, aiming at both CO and machine learning researchers. To this end, we thoroughly introduce CO, the various machine learning regimes, and GNNs. Most importantly, we give a comprehensive, structured overview of recent applications of GNNs in the CO context. 
We discuss challenges arising from the use of GNNs and future work. Our contributions can be summarized as follows:
\begin{enumerate}
	\item We provide a complete, structured overview of the application of GNNs to the CO setting for both heuristic and exact algorithms.
	\item We survey recent progress in using GNN-based end-to-end algorithmic reasoners.
	\item We highlight the shortcomings of GNNs in the context of CO and provide guidelines and recommendations on how to tackle them.
	\item We provide a list of open research directions to stimulate future research.
\end{enumerate}

\subsection{Related Work}
In the following, we briefly review key papers and survey efforts involving GNNs and machine learning for CO.

\paragraph{GNNs} Graph neural networks~\citep{Gil+2017,Sca+2009} have recently (re-)emerged as the leading machine learning method for graph-structured inputs. Notable instances of this architecture include, e.g.,~\cite{Duv+2015,Ham+2017,Vel+2018}, and the spectral approaches proposed by, e.g.,~\citet{Bru+2014,Defferrard2016,Kip+2017,Mon+2017}---all of which descend from early work of~\citet{Kir+1995,Spe+1997,Mer+2005,Sca+2009}. Aligned with the field's recent rise in popularity, there exists a plethora of surveys on recent advances in GNN techniques. Some of the most recent ones include~\cite{Cha+2020,Wu+2019,Zho+2018}.

\paragraph{Continuous Formulations}The discrete nature of CO problems makes standard continuous optimization tools unavailable, such as first- and second-order gradient methods. However, many problems admit alternative reformulations as non-convex continuous optimization problems over graphs. Such problems include graph partitioning, maximum cut, minimum vertex cover, maximum independent set, and maximum clique problems. Some early work at the intersection of machine learning and combinatorial optimization involves reinterpreting these continuous optimization problems as energy-based training of Hopfield neural networks, or self-organizing maps, such as in the work of \cite{hopfield1985neural}, \cite{durbin1987analogue}, \cite{ramanujam1995mapping} and \cite{gold1996softmax}. Although not using GNNs, these works use graphs as a central object. They can be seen as foreshadowing various GNN-based differentiable proxy loss approaches that we summarize in Section \ref{subsub:lhs}.

\paragraph{Surveys} The seminal survey of~\citet{smith1999neural} centers around the use of popular neural network architectures of the time, namely Hopfield Networks and Self-Organizing Maps, as a basis for combinatorial heuristics, as described in the previous section. It is worth noting that such architectures were mostly used for a single instance at a time, rather than being trained over a set of training instances; this may explain their limited success at the time. \citet{Bengio2018} give a high-level overview of machine learning methods for CO, with no special focus on graph-structured input, while \citet{Lod+2017} focus on machine learning for branching in the context of mixed-integer programming. Concurrently to our work, \citet{kotary2021end} have categorized various approaches for machine learning in CO, focusing primarily on end-to-end learning setups and paradigms, making representation learning---and GNNs in particular---a secondary topic. Moreover, the surveys by \cite{Maz+2020,Yan+2020} focus on using reinforcement learning for CO. The survey of~\cite{Ves+2020} deals with machine learning for network problems arising in telecommunications, focusing on non-exact methods and not including recent progress. Finally,~\cite{Lam+2020} give a high-level overview of the application of GNNs in various reasoning tasks, missing out on the most recent developments, e.g., the algorithmic reasoning direction that we study in detail here.

\subsection{Outline}
We start by giving the necessary background on CO and relevant optimization frameworks, machine learning, and GNNs; see~\cref{prelim}. In~\cref{cognn}, we review recent research using GNNs in the CO context. Specifically, in~\cref{primal}, we survey works aiming at finding primal solutions, i.e., high-quality feasible solutions to CO problems, while~\cref{bb} gives an overview of works aiming at enhancing dual methods, i.e., proving the optimality of solutions. Going beyond that,~\cref{algo} reviews recent research trying to facilitate algorithmic reasoning behavior in GNNs, as well as applying GNNs as raw-input combinatorial optimizers. Finally,~\cref{limits} discusses the limits of current approaches and offers a list of research directions, with the aim of stimulating future research.

\section{Preliminaries}\label{prelim}

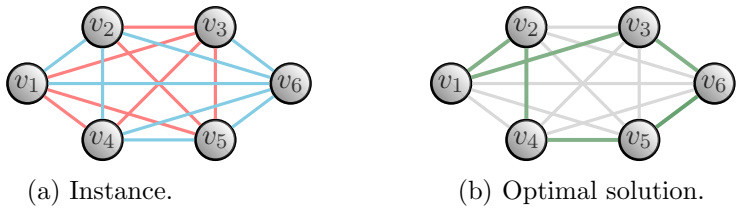
\begin{figure}
	\centering

	\begin{subfigure}[t]{1in}\centering
		\begin{tikzpicture}
			\def\bend{15}
			\def\opac{0.2}

			\tikzset{line/.style={draw,line width=1.2pt}}
			\tikzset{arrow/.style={line,->,>=stealth}}
			\tikzset{snake/.style={arrow,line width=1.3pt,decorate,decoration={snake,amplitude=1,segment length=6,post length=7}}}
			\tikzset{box/.style={dash pattern=on 5pt off 2pt,inner sep=5pt,rounded corners=3pt}}
			\tikzset{node/.style={shape=circle,inner sep=0pt,minimum width=15pt,line width=1pt}}
			\tikzset{light/.style={shading=axis,left color=white,right color=black,shading angle=-45}}

			\tikzset{pics/bar/.style args={#1/#2/#3}{code={%
								\node[inner sep=0pt,minimum height=0.3cm] (#1) at (0,-0.15) {};

								\draw[fill=blue,blue] (-0.15,-0.2) rectangle (-0.075,0.25*#2-0.2);
								\draw[fill=orange,orange] (-0.025,-0.2) rectangle (0.05,0.25*#3-0.2);

								\begin{scope}[transparency group,opacity=\opac]
									\draw[fill=red,light] (-0.15,-0.2) rectangle (-0.075,0.25*#2-0.2);
									\draw[fill=orange,light] (-0.025,-0.2) rectangle (0.05,0.25*#3-0.2);
								\end{scope}

								\draw[arrow,line width=1pt] (-0.2,-0.2) -- (0.2,-0.2);
								\draw[arrow,line width=1pt] (-0.2,-0.21) -- (-0.2,0.2);
							}}}

			\node[line,node] (x1) at (0.5, -0.75) {$v_1$};
			\node[line,node] (x2) at (1.5, 0) {$v_2$};
			\node[line,node] (x3) at (3, 0) {$v_3$};
			\node[line,node] (x4) at (1.5, -1.5) {$v_4$};
			\node[line,node] (x5) at (3, -1.5) {$v_5$};
			\node[line,node] (x6) at (4.0, -0.75) {$v_6$};

			\begin{scope}[transparency group,opacity=\opac]
				\node[line,node,light] at (x1) {};
				\node[line,node,light] at (x2) {};
				\node[line,node,light] at (x3) {};
				\node[line,node,light] at (x4) {};
				\node[line,node,light] at (x5) {};
				\node[line,node,light] at (x6) {};

			\end{scope}

			\path[line, red!50] (x1) to (x3);
			\path[line, red!50] (x1) to (x4);
			\path[line, red!50] (x1) to (x5);
			\path[line, red!50] (x2) to (x3);
			\path[line, red!50] (x2) to (x5);
			\path[line, red!50] (x3) to (x4);
			\path[line, red!50] (x3) to (x5);

			\path[line, blue!50] (x1) to (x6);
			\path[line, blue!50] (x1) to (x2);
			\path[line, blue!50] (x2) to (x4);
			\path[line, blue!50] (x3) to (x6);
			\path[line, blue!50] (x2) to (x6);
			\path[line, blue!50] (x4) to (x5);
			\path[line, blue!50] (x4) to (x6);
			\path[line, blue!50] (x5) to (x6);
		\end{tikzpicture}
		\caption{Instance.}
	\end{subfigure}\hspace{70pt}
	\begin{subfigure}[t]{2in}\centering
		\begin{tikzpicture}
			\def\bend{15}
			\def\opac{0.2}

			\tikzset{line/.style={draw,line width=1.2pt}}
			\tikzset{arrow/.style={line,->,>=stealth}}
			\tikzset{snake/.style={arrow,line width=1.3pt,decorate,decoration={snake,amplitude=1,segment length=6,post length=7}}}
			\tikzset{box/.style={dash pattern=on 5pt off 2pt,inner sep=5pt,rounded corners=3pt}}
			\tikzset{node/.style={shape=circle,inner sep=0pt,minimum width=15pt,line width=1pt}}
			\tikzset{light/.style={shading=axis,left color=white,right color=black,shading angle=-45}}

			\tikzset{pics/bar/.style args={#1/#2/#3}{code={%
								\node[inner sep=0pt,minimum height=0.3cm] (#1) at (0,-0.15) {};

								\draw[fill=blue,blue] (-0.15,-0.2) rectangle (-0.075,0.25*#2-0.2);
								\draw[fill=orange,orange] (-0.025,-0.2) rectangle (0.05,0.25*#3-0.2);

								\begin{scope}[transparency group,opacity=\opac]
									\draw[fill=red,light] (-0.15,-0.2) rectangle (-0.075,0.25*#2-0.2);
									\draw[fill=orange,light] (-0.025,-0.2) rectangle (0.05,0.25*#3-0.2);
								\end{scope}

								\draw[arrow,line width=1pt] (-0.2,-0.2) -- (0.2,-0.2);
								\draw[arrow,line width=1pt] (-0.2,-0.21) -- (-0.2,0.2);
							}}}

			\node[line,node] (x1) at (0.5, -0.75) {$v_1$};
			\node[line,node] (x2) at (1.5, 0) {$v_2$};
			\node[line,node] (x3) at (3, 0) {$v_3$};
			\node[line,node] (x4) at (1.5, -1.5) {$v_4$};
			\node[line,node] (x5) at (3, -1.5) {$v_5$};
			\node[line,node] (x6) at (4.0, -0.75) {$v_6$};

			\begin{scope}[transparency group,opacity=\opac]
				\node[line,node,light] at (x1) {};
				\node[line,node,light] at (x2) {};
				\node[line,node,light] at (x3) {};
				\node[line,node,light] at (x4) {};
				\node[line,node,light] at (x5) {};
				\node[line,node,light] at (x6) {};

			\end{scope}

			\path[line, black!15] (x4) to (x6);

			\path[line, black!15] (x1) to (x4);
			\path[line, black!15] (x1) to (x5);
			\path[line, black!15] (x1) to (x6);

			\path[line, black!15] (x2) to (x3);
			\path[line, black!15] (x2) to (x5);
			\path[line, black!15] (x2) to (x6);

			\path[line, black!15] (x3) to (x4);
			\path[line, black!15] (x3) to (x5);
			\path[line, green!50,line width=1.5pt] (x3) to (x6);
			\path[line, green!50,line width=1.5pt] (x2) to (x4);

			\path[line, green!50,line width=1.5pt] (x4) to (x5);
			\path[line, green!50,line width=1.5pt] (x1) to (x2);
			\path[line, green!50,line width=1.5pt] (x1) to (x3);

			\path[line, green!60,line width=1.5pt] (x5) to (x6);
		\end{tikzpicture}
		\caption{Optimal solution.}

	\end{subfigure}

	\caption{A complete graph with edge labels (blue and red) and its optimal solution for the TSP (in green). Blue edges have a cost of $1$ and red edges a cost of $2$.}\label{fig:tsp}

\end{figure}

Here, we introduce notation and give the necessary formal background on combinatorial optimization, the different machine learning regimes, and GNNs.

\subsection{Notation}

Let $[n] = \{ 1, \dotsc, n \} \subset \NN$ for $n \geq 1$, and let $\{\!\!\{ \dots\}\!\!\}$ denote a multiset. For a (finite) set $S$, we denote its \new{power set} as $2^S$.

A \new{graph} $G$ is a pair $(V,E)$ with a \emph{finite} set of
\new{nodes} $V$ and a set of \new{edges} $E \subseteq V\times V$.
We denote the set of nodes and the set
of edges of $G$ by $V(G)$ and $E(G)$, respectively.
A \new{labeled graph} $G$ is a triplet
$(V,E,l)$ with a label function $l \colon V(G) \cup E(G) \to \Sigma$,
where $\Sigma$ is some finite alphabet. Then, $l(x)$ is a
\new{label} of $x$, for $x$ in $V(G) \cup E(G)$. Note that $x$ here can be either a node or an edge.
The \new{neighborhood}
of $v$ in $V(G)$ is denoted by $N(v) = \{ u \in V(G) \mid (v, u) \in E(G) \}$.
A \new{tree} is a connected graph without
cycles.

We say that two graphs $G$ and $H$
are \new{isomorphic} if there exists an edge-preserving bijection
$\varphi \colon V(G) \to V(H)$, i.e., $(u,v)$ is in $E(G)$ if and only if
$(\varphi(u),\varphi(v))$ is in $E(H)$. For labeled graphs, we further require that
$l(v) = l(\varphi(v))$ for $v$ in $V(G)$ and $l((u,v)) = l((\varphi(u), \varphi(v)))$ for $(u,v)$ in $E(G)$.

\subsection{Combinatorial Optimization}
CO deals with problems that involve optimizing a cost (or objective) function by selecting a subset from a finite set, with the latter encoding constraints on the solution space. Formally, we define an instance of a \new{combinatorial optimization problem} as follows.
\begin{definition}[Combinatorial optimization instance]
	An instance of a \emph{combinatorial optimization problem} is a tuple $(\Omega, F, w)$, where $\Omega$ is a \emph{finite} set, $F \subseteq 2^\Omega$ is the set of \emph{feasible} solutions, $c \colon 2^\Omega \to \bbR$ is a \emph{cost function} with $c(S) = \sum_{\omega \in S} w(\omega)$ for $S$ in $F$. 
\end{definition}
Consequently, CO deals with selecting an element $S^*$ (\new{optimal solution}) in $F$ that minimizes $c$ over the feasible set $F$.\footnote{Without loss of generality, we choose minimization instead of maximization.} The corresponding \new{decision problem} asks if there exists an element in the feasible set such that its cost is smaller than or equal to a given value, i.e., whether there exists  $S$ in $F$ such that $c(S) \leq k$ (i.e., we require a \textsc{Yes}/\textsc{No} answer).

The TSP is a well-known CO problem aiming at finding a cycle along the edges of a graph with minimal cost that visits each node exactly once; see~\cref{fig:tsp} for an illustration of an instance of the TSP problem and its optimal solution. The corresponding decision problem asks whether there exists a cycle along the edges of a graph with cost $\leq k$ that visits each node exactly once.

\begin{example}[Traveling Salesperson Problem]\hspace{1pt}\\\label{tspex}\textit{Input:} A complete directed graph $G$, i.e., $E(G) = \{ (u,v) \mid u,v \in V(G) \}$, with edge costs $w\colon E(G) \to \mathbb{R}.$\\ 
	\textit{Output:} A permutation of the nodes $\sigma \colon \{ 0, \dots, n-1 \} \to V$ such that
	\begin{equation*}
		\sum_{i=0}^{n-1} w \big((\sigma{(i)} ,\sigma{((i+1) \bmod n)} \big)
	\end{equation*}
	is minimal over all permutations, where $n = |V|$.
\end{example}
Due to their discrete nature, many classes or sets of combinatorial decision problems arising in practice, e.g., TSP or other vehicle routing problems, are \textsf{NP}-hard\footnote{In complexity theory, one refers to the optimization version of such difficult problems as \textsf{NP}-hard, while their decision version is generally \textsf{NP}-complete.}~\citep{Kor+2012}, and hence likely {intractable} in the worst-case sense. However, instances are routinely solved in practice by formulating them as \emph{integer linear optimization problems} or \emph{integer linear programs} (ILPs), \emph{constrained problems}, or as \emph{satisfiability problems} (SAT) and utilizing well-engineered algorithms (and associated solvers) for these problems, e.g., branch-and-cut algorithms in the case of ILPs; see the next section for details.

\subsection{General Optimization Frameworks: ILPs, SAT, and Constrained Problems}

In the following, we describe common modeling and algorithmic frameworks for CO problems. More precisely, the next three sections describe the modeling approaches, namely, integer programming, SAT, and constraint satisfaction/optimization. Finally, Section \ref{sec:algo} partitions the algorithmic frameworks into three categories.

\subsubsection{Integer linear programs and mixed-integer programs}
\label{sec:ilp}

First, we start by defining a \emph{linear program} or \emph{linear optimization problem}. A linear program aims at optimizing a linear cost function over a feasible set described as the intersection of finitely many half-spaces, i.e., a polyhedron. Formally, we define an instance of a linear program as follows.
\begin{definition}[Linear programming instance]
	An instance of a \emph{linear program} (LP) is a tuple $(A,b,c)$, where $A$ is a matrix in $\bbR^{m \times n}$, and $b$ and $c$ are vectors in $\bbR^{m}$ and $\bbR^{n}$, respectively. 
\end{definition}
The associated optimization problem asks to minimize a linear objective over a polyhedron.\footnote{In the above definition, we assumed that the LP is feasible, i.e., $X \neq \emptyset$, and that a finite minimum value exists. In what
	follows, we assume that both conditions are always fulfilled.}
That is, we aim at finding a vector $x$ in $\bbR^{n}$ that minimizes $c^T x$ over the \emph{feasible set} 
\begin{equation*}
	X = \{ x \in \bbR^n \mid A_j x \leq b_j \text{ for } j \in [m]  \text{ and }  x_i \geq 0 \text{ for } i \in [n]  \}.
\end{equation*} 
In practice, LPs are solved using the Simplex method or polynomial-time interior-point methods~\citep{Ber+2007}. Due to their continuous nature, LPs cannot encode the feasible set of a CO problem. Hence, we extend LPs by adding \new{integrality constraints}, i.e., requiring that the value assigned to each variable is an integer. Consequently, we aim to find the vector $x$ in $\bbZ^{n}$ that minimizes $c^T x$ over the feasible set
\begin{equation*}
	X = \{ x \in \bbZ^n \mid A_j x \leq b_j  \text{ for } j \in [m] ,  x_i \geq 0 \text{ and } x_i \in \mathbb{Z} \text{ for } i \in [n]  \}.
\end{equation*}
Such integer linear optimization problems are solved by tree search algorithms, e.g., branch-and-bound algorithms, see~\cref{sec:algo} for details. By dropping the integrality constraints, we again obtain an instance of an LP, which we call \emph{relaxation}. Solving the LP relaxation of an ILP provides a valid lower bound on the optimal solution of the problem, i.e., an optimistic approximation, and the quality of such an approximation is largely responsible of the effectiveness of the search scheme.
\begin{example}
	\label{example:tsp}
	We provide an ILP that encodes all feasible solutions of the TSP, and, due to the objective function, selects the optimal one. Essentially, it encodes the order of the nodes or cities within its variables. Thereto, let
	$$x_{ij} = \begin{cases} 1 & \text{if the cycle goes from city } i \text{ to city } j, \\ 0 & \text{otherwise,} \end{cases}$$
	and let $w_{ij} > 0$ be the cost or distance of traveling from city $i$ to city $j$, $i\neq j$. Then, the TSP can be written as the following ILP:\footnote{Technically, the presented TSP model is for the \emph{asymmetric} version, where the costs $w_{ij}$ and $w_{ji}$ might be different. Such a TSP version is represented in a directed graph. Instead, the version in Figure \ref{fig:tsp} is \emph{symmetric}, i.e., $w_{ij}=W_{ji}$, and it is represented on an undirected graph.}
	\begin{align*}
		\min                 & \,\,\sum_{i=1}^n \sum_{j\ne i,j=1}^n w_{ij}x_{ij}          &  &                                       \\
		\mathrm{subject\ to} & \sum_{i=1,i\ne j}^n x_{ij} = 1                             &  & j \in [n],                            \\
		                     & \sum_{j=1,j\ne i}^n x_{ij} = 1                             &  & i \in [n],                            \\
		                     & \,\,\sum_{i \in Q}{\sum_{j \not \in Q}{x_{ij}}} \geq 1 &  & \forall Q \subsetneq [n], |Q| \geq 2. \\
	\end{align*}
	The first two constraints encode that each city should have exactly one in-going and out-going edge, respectively. The last constraint makes sure that all cities are within the same tour, i.e., there exist no sub-tours (thus, the returned solution is not a collection of smaller tours).
\end{example}

In practice, one often faces problems consisting of a mix of integer and continuous variables. These are commonly known as \emph{mixed-integer programs} (MIPs). Formally, given an integer $p > 0$, MIPs aim at finding a vector $x$ in $\bbR^{n}$ that minimizes $c^T x$ over the \emph{feasible set} 
\begin{equation*}
	X = \{ x \in \bbR^n \mid A_j x \leq b_j \text{ for } j \in [m], x_i \geq 0 \text{ for } i \in [n], \text{ and } x \in \bbZ^{p} \times \bbR^{n-p}  \}.
\end{equation*} 
Here, $n$ is the number of variables we are optimizing, out of which $p$ are required to be integers.

\subsubsection{SAT}
The \emph{Boolean satisfiability problem} (SAT) asks, given a Boolean formula or propositional logic formula, if there exists a variable assignment (assign \emph{true} or \emph{false} to variables) such that the formula evaluates to \emph{true}. Hence, formally we can define it as follows.
\begin{definition}[SAT]\hspace{1pt}\\
	\textit{Input:} A propositional logic formula $\varphi$ with variable set $V$.\\
	\textit{Output:} \textsc{Yes}, if there exists a variable assignment $A \colon V \to \{ \text{true}, \text{false} \}$ such that the formula $\varphi$ evaluates to $\text{true}$; \textsc{No}, otherwise.
\end{definition}
The SAT problem was the first one to be shown to be \textsf{NP}-complete \citep{cook1971complexity}, however, modern solvers routinely solve industrial-scale instances in practice~\citep{Pra+2005}.
Despite the simplicity of its formalization, SAT has many practical applications, such as hardware verification 
\citep{clarke2003sat,gupta2006sat}, configuration management \citep{mancinelli2006managing,tucker2007opium}, or planning \citep{behnke2018totsat}. A realistic case study of SAT is illustrated in Example \ref{ex:sat}. 

\begin{example}\hspace{1pt} \label{ex:sat} Let us consider the problem of installing a new (software) package $P$ on a system, where the installation is subject to dependency and conflict constraints.
The goal is to determine which packages must be installed on the system such that, the package $P$ is installed in the system, 
the dependencies of all the installed packages are satisfied, 
and there are no conflicts among the installed packages.
This problem can be conveniently modeled as a SAT problem \citep{tucker2007opium}. 
Formally, let $I$ be the set of packages involved in the installation, and let $x_i$ be a Boolean variable stating if the package $i$ in $I$ is installed. The constraints are encoded as follows: 
(1) $x_P$, it ensures that the target package $P$ is installed, 
(2) $x_A \to x_B$, it ensures that the package $A$ can be installed only if package $B$ is also installed (dependency between packages),
(3) $\neg x_A \lor \neg x_B$, it ensures that packages $A$ and $B$ cannot be installed together (conflict between packages).
Assuming that $C^d_1,\dots,C^d_n$ are the dependency constraints, and that $C^c_1,\dots,C^c_m$
are the conflict constraints, the Boolean formula to resolve corresponds to the logical conjunction of all the constraints, i.e.,
\begin{equation*}
\varphi = x_P \land C^d_1 \land \dots \land C^d_n \land C^c_1 \land \dots \land C^c_m.
\end{equation*}
Many variants can be inferred from this formalization, such as integrating dependencies among more packages or finding the minimum set of packages that must be installed.
\end{example}

\subsubsection{Constraint satisfaction and constraint optimization problems}

This section presents both  \emph{constraint satisfaction problems} and \emph{constraint optimization problems}, the most generic way to formalize CO problems.
Formally, an instance of a \new{constraint satisfaction problem} is defined as follows.

\begin{definition}[Constraint satisfaction problem instance]
	An instance of a constraint satisfaction problem (CSP) is a tuple $\big( X, D(X), C \big)$, where $X$ is the set of variables, $D(X)$ is the set of domains of the variables, and $C$ is the set of constraints that restrict assignments of values to variables.
	A solution is an assignment of values from $D$ to  $X$ that satisfies all the constraints of $C$.
\end{definition}

A natural extension of CSPs are \textit{constrained optimization problems}, i.e., CSPs that also have an objective function. The goal becomes finding a feasible assignment that minimizes the objective function. The main difference with the previous optimization frameworks is that constrained optimization problems do not require underlying assumptions on the variables, constraints, and objective functions. Unlike MIPs, non-linear objectives and constraints are applicable within this framework. For instance, a TSP model is presented next.

\begin{example}
\label{ex:cp}
Given a configuration with $n$ cities and a  weight matrix $w$ in $\mathbb{R}^{n \times n}$, the TSP can be modeled using $n$ variables $x_i$ over the domains $D(x_i) \colon [n]$. 
	Variable $x_i$ indicates the  $i$-th city to be visited. The objective function and constraints read as
	\label{ex:cp-tsp}
	\begin{align*}
		\min                 & \,\, w_{x_{n},x_{1}} + \sum_{i=1}^{n-1} w_{x_i,x_{i+1}} &  & \\
		\mathrm{subject\ to} & \;\; \textsc{allDifferent}(x_1,\dots,x_n),              &  &
	\end{align*}
	where \textsc{allDifferent}$(X)$  enforces that each variable from $X$ takes a different value~\citep{regin1994filtering}, and the entries of the weight matrix $w$ are indexed using variables.
	This model enforces each city to have another city as a successor and sums up the distances between each pair of consecutive cities along the cycle.
\end{example}

As shown in the above example, constrained problems can model arbitrary constraints and objective functions.
This generality makes it possible to use general-purpose solving methods such as \emph{local search} or \emph{constraint programming} (see next section).
In addition to their convenience on the modeling side, the high-level constraints, generally referred to as \textit{global constraints}, are also useful on the solving side~\citep{regin2004global}.
They enable the design of efficient algorithms dedicated to prune the search space.
Leveraging the pruning ability of global constraints is a fundamental component of a constraint programming solver as explained below.

\subsubsection{Solving CO problems}
\label{sec:algo}

Major algorithmic frameworks---whose components and tasks have been recently considered through the GNN lens---will be discussed when necessary in the core of the survey. However, in this section, we briefly distinguish three algorithmic categories.\footnote{
We refer to~\cite{festa2014brief} for additional details on the classification of algorithms for CO.}

\paragraph{Exact methods} ILP models are generally solved to proven optimality (or a proof of infeasibility) by variations of the \emph{branch-and-bound algorithm} \citep{LD60, lodi2010mixed}. Essentially, the algorithm is an iterative divide-and-conquer method that
\begin{enumerate}
    \item solves LP relaxations (see \cref{sec:ilp}), 
    \item \label{en:2} improves them through valid inequalities (or \emph{cutting planes}), and 
    \item guarantees to find an optimal solution through implicit enumeration performed by \emph{branching}, see Figure \ref{fig:BnB-mdp}. 
\end{enumerate}
As anticipated in Section \ref{sec:ilp}, the quality of the LP relaxation plays a fundamental role in the effectiveness of the above scheme. Thus, step \ref{en:2} above is particularly important, especially at the beginning of the search. The above scheme is called \emph{branch and cut}.
If the CO problem is not directly modeled by integer programming techniques, then combinatorial versions of the branch-and-bound framework are devised, i.e., featuring relaxations different from the LP one, specifically associated with the structure of the CO problem at hand. 

Specifically designed as an exact method for constrained satisfaction and optimization problems, \emph{constraint programming} (CP) \citep{rossi2006handbook} is a general framework proposing simple algorithmic solutions also within the divide-and-conquer scheme. It is a complete approach, meaning it is possible to {prove} the optimality of the solutions found. The solving process consists of a complete enumeration of all possible variable assignments until the best solution has been found. To cope with the implied (exponentially) large search trees, one utilizes a mechanism called \textit{propagation}, which reduces the number of possibilities. Here, the propagation of the constraint $c$ removes values from domains violated by $c$. This process is repeated at each domain change and for each constraint until no value exists anymore. The efficiency of a CP solver relies heavily on the quality of its propagators. 
Example \ref{ex:cp} introduced the well-known \textsc{allDifferent} constraint. Its propagator~\citep{regin1994filtering} is based on maximum matching algorithms in a graph.
There exist many other global constraints that are available in the literature, such as \textsc{Element} constraint, allowing indexation with variables,
or \textsc{Circuit} constraint, enforcing a set of variables to create a valid circuit. At the time of writing, the global constraints catalog reports more than 400 different global constraints \citep{beldiceanu2005global}.
The CP search commonly proceeds in a depth-first fashion, together with branch-and-bound. For each  feasible solution found, the solver adds a constraint, ensuring that the following solution has to be better than the current one. Upon finding an infeasible solution, the search backtracks to the previous decision. With this procedure, and provided that the whole search space has been explored, the final solution found is then guaranteed to be optimal. 

Finally, although initially designed for solving decision problems, SAT solvers can also be used for combinatorial optimization.
One way to do that is to specify objectives through soft constraints. 
The objective turns to satisfy as many soft constraints as possible in a solution. 
Another option is to add a repertoire of common objective functions in the solver and invoke
the specialized optimization module when required. Modern SAT solvers such as Z3 generally support both options \citep{moura2008z3}.

\paragraph{Local search and metaheuristics} Local search~\citep{potvin2018handbook} is another algorithmic framework that is commonly used to solve general, large-scale CO problems. Local search only partially explores the solution space in a perturbative fashion and is thus an incomplete approach that does not provide an optimality guarantee on the solution it returns. In its simplest form, the search starts from a candidate solution $s$ and iteratively explores the solution space by selecting a neighboring solution until no improvement occurs. Here, the \emph{neighborhood} of a solution is the set of solutions obtained by making some modifications to the solution $s$. In practice, 
local search algorithms are improved through \textit{metaheuristic} concepts \citep{glover2006handbook}, 
leading to algorithms like \textit{simulated annealing} \citep{van1987simulated,delahaye2019simulated},
\textit{tabu search} \citep{glover1998tabu,laguna2018tabu}, \textit{genetic algorithms} \citep{kramer2017genetic}, 
\textit{variable neighborhood search} \citep{mladenovic1997variable,hansen2019variable}, all of which are designed to help escape \textit{local minima}.

\paragraph{Approximation algorithms} The class of \emph{approximation algorithms} \citep{Vaz+2010} is designed to produce, typically in polynomial time, feasible solutions for CO problems. Unlike local search and metaheuristics,
the value of those feasible solutions is guaranteed to be within a certain bound from the optimal one. 
Notable examples of approximation algorithms are \textit{polynomial-time approximation schemes} (PTAS) 
that provide a solution that is within a factor $1 + \epsilon$ (with $\epsilon > 0$ being an input for the algorithm) of being optimal (e.g., \citet{arora1996polynomial} for the TSP),
or \textit{fully polynomial-time approximation schemes} (FPTAS), where additional conditions on the running time of the algorithm \citep{ausiello2012complexity} are imposed.

\subsection{Machine Learning}
\label{sec:mlparadigms}

In this section, we give a short and concise overview of machine learning. We cover the three main branches of the field, i.e., \new{supervised learning}, \new{unsupervised learning}, and \new{reinforcement learning}. For details, see~\cite{Moh+2012,Shalev-Shwartz2014}. Moreover, we introduce \new{imitation learning} that is of high relevance to CO.

\paragraph{Supervised learning} Given a finite training set, i.e., a set of examples (e.g., graphs) together with target values (e.g., real values in the case of regression), supervised learning tries to adapt the parameters of a model (e.g., a neural network) based on the examples and targets. The adaptation of the parameters is achieved by minimizing a loss function that measures how well the chosen parameters align with the target values.
Formally, let $\mathcal{X}$ be the set of possible \new{examples} and let $\mathcal{Y}$ be the set of possible \new{target values}. We assume that the pairs in $\mathcal{X} \times \mathcal{Y}$ are independently and identically distributed (i.i.d.) with respect to a fixed but unknown distribution $\mathcal{D}$. Moreover, we assume that there exists a \new{target concept} $c\colon \mathcal{X} \to \mathcal{Y}$ that maps each example to its target value. Given a sample $S = ((s_1,c(s_1)), \dots, (s_m,c(s_m)))$ drawn i.i.d. from $\mathcal{D}$, the aim of supervised machine learning is to select a \new{hypothesis} $h \colon \mathcal{X} \to \mathcal{Y}$ from the set of possible hypotheses by minimizing the \new{empirical error}
$\widehat{R}(h) = \frac{1}{m} \sum_{i=1}^m \ell(h(s_i), c(s_i)),$ where $\ell \colon \mathcal{X} \times \mathcal{Y}\to \RR$ is the loss function. To avoid overfitting to the given samples, we add a \new{regularization penalty} $\Omega \colon H \to \bbR$ to the empirical error. Examples of supervised machine learning methods include neural networks, support vector machines, and boosting.

\paragraph{Unsupervised learning} Unlike supervised learning, there is no training set in the unsupervised case, i.e., no target values are available. Accordingly, unsupervised learning aims to capture representative characteristics of the data ({features}) by minimizing an unsupervised loss function, $l \colon \mathcal{X} \to \RR$. In this case, the loss function only directly depends on the input samples $s_i$, as no labels are provided upfront. Examples of unsupervised machine learning methods include autoencoders, clustering, and principal component analysis.

\paragraph{Reinforcement learning (RL)} Similarly to unsupervised learning, reinforcement learning does not rely on a labeled training set. Instead, an \new{agent} explores an environment, e.g., a graph,
by taking \emph{actions}. To guide the agent in its exploration, it receives two types of feedback, its current \new{state}, and a \new{reward}, usually a real-valued scalar, indicating how well it achieved its goal so far. The RL agent aims to maximize the cumulative reward it receives by determining the best actions.
Formally, let $( S, A, T, R)$ be a tuple representing a \textit{Markov decision process} (MDP). Here, $S$ is the set of \new{states} in the environment and $A$ is the set of \new{actions} that the agent can do. The function $T \colon S \times S \times A \to [0,1]$ is the \new{transition probability function} giving the probability, $T(s, s', a)$, of transitioning from $s$ to $s'$ if action $a$ is performed, such that $\sum_{s'}^S T(s, s', a) = 1$ for all $s$ in $S$ and  $a$ in $A$.
Finally, $R  \colon S \times A \to \mathbb{R}$ is the \new{reward function} of taking an action from a specific state.
An agent's behavior is defined by a \new{policy} $\pi  \colon S \times A \to [0,1]$, describing the probability of taking an action from a given state.
From an initial state $s_1$, the agent performs actions, yielding a sequence of states until reaching a terminal state, $s_{\Theta}$. Such a sequence $s_1\dots,s_\Theta$ is referred to as an \new{episode}. An agent's goal is to learn a policy maximizing the cumulative sum of rewards, eventually discounted by a value $\gamma$ in $[0,1]$, during an episode, i.e.,  $\sum_{k=1}^{\Theta}  \gamma^{k} R(s_k, a_k)$ is maximized.  While such a learning setting is very general, the number of combinations increases exponentially with the number of states and actions, quickly making the problem intractable.
Excluding hybrid approaches, e.g., RL with Monte Carlo tree search~\citep{browne2012survey} and model-based approaches~\citep{polydoros2017survey}, there exist two kinds of reinforcement learning algorithms, \new{value-based methods}, aiming to learn a function characterizing the goodness of each action,
and \new{policy-based methods}, aiming to learn the policy directly.

\paragraph{Imitation learning}

Imitation learning~\citep{ross2013interactive} attempts to solve sequential decision-making problems by {imitating} another (``expert'') policy rather than relying on rewards for feedback as done in RL. This makes imitation learning attractive for CO because, for many control problems, one can devise rules that make excellent decisions but are not practical because of computational cost or because they cheat by using information that would not be available at solving time.
Imitation learning algorithms can be offline or online. When offline, examples of expert behavior are collected beforehand, and the student policy's training is done subsequently. In this scenario, training is simply a form of supervised learning. When online, however, the training occurs while interacting with the environment, usually by querying the expert for advice when encountering new states. Online algorithms can be further subdivided into on-policy and off-policy algorithms. In on-policy algorithms, the distribution of states from which examples of expert actions were collected matches the student policy's stationary distribution to be updated. In off-policy algorithms, there is a mismatch between the distribution of states from which the expert was queried and the distribution of states the student policy is likely to encounter. Some off-policy algorithms attempt to correct this mismatch accordingly.

\subsection{Graph Neural Networks}

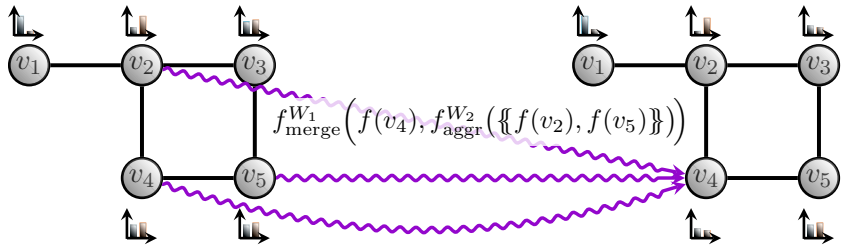
\begin{figure}
	\centering
	\begin{tikzpicture}

		\def\bend{15}
		\def\opac{0.2}

		\tikzset{line/.style={draw,line width=1.5pt}}
		\tikzset{arrow/.style={line,->,>=stealth}}
		\tikzset{snake/.style={arrow,line width=1.3pt,decorate,decoration={snake,amplitude=1,segment length=6,post length=7}}}
		\tikzset{box/.style={dash pattern=on 5pt off 2pt,inner sep=5pt,rounded corners=3pt}}
		\tikzset{node/.style={shape=circle,inner sep=0pt,minimum width=15pt,line width=1pt}}
		\tikzset{light/.style={shading=axis,left color=white,right color=black,shading angle=-45}}

		\tikzset{pics/bar/.style args={#1/#2/#3}{code={%
							\node[inner sep=0pt,minimum height=0.3cm] (#1) at (0,-0.15) {};

							\draw[fill=blue,blue] (-0.15,-0.2) rectangle (-0.075,0.25*#2-0.2);
							\draw[fill=orange,orange] (-0.025,-0.2) rectangle (0.05,0.25*#3-0.2);

							\begin{scope}[transparency group,opacity=\opac]
								\draw[fill=red,light] (-0.15,-0.2) rectangle (-0.075,0.25*#2-0.2);
								\draw[fill=orange,light] (-0.025,-0.2) rectangle (0.05,0.25*#3-0.2);
							\end{scope}

							\draw[arrow,line width=1pt] (-0.2,-0.2) -- (0.2,-0.2);
							\draw[arrow,line width=1pt] (-0.2,-0.21) -- (-0.2,0.2);
						}}}

		\node[line,node] (x1) at (0, 0) {$v_1$};
		\node[line,node] (x2) at (1.5, 0) {$v_2$};
		\node[line,node] (x3) at (3, 0) {$v_3$};
		\node[line,node] (x4) at (1.5, -1.5) {$v_4$};
		\node[line,node] (x5) at (3, -1.5) {$v_5$};

		\pic[] at (0,0.6) {bar=x1-bar/1.0/0.2};
		\pic[] at (1.5,0.6) {bar=x2-bar/0.4/1.0};
		\pic[] at (3,0.6) {bar=x3-bar/0.7/0.8};
		\pic[] at (1.5,-2.1) {bar=x4-bar/0.7/0.8};
		\pic[] at (3,-2.1) {bar=x5-bar/0.7/0.8};

		\begin{scope}[transparency group,opacity=\opac]
			\node[line,node,light] at (x1) {};
			\node[line,node,light] at (x2) {};
			\node[line,node,light] at (x3) {};
			\node[line,node,light] at (x4) {};
			\node[line,node,light] at (x5) {};
		\end{scope}

		\path[line] (x1) to (x2);
		\path[line] (x2) to (x3);
		\path[line] (x2) to (x4);
		\path[line] (x3) to (x5);
		\path[line] (x4) to (x5);

		\node[line,node] (y1) at (7.5, 0) {$v_1$};
		\node[line,node] (y2) at (9, 0) {$v_2$};
		\node[line,node] (y3) at (10.5, 0) {$v_3$};
		\node[line,node] (y4) at (9, -1.5) {$v_4$};
		\node[line,node] (y5) at (10.5, -1.5) {$v_5$};

		\begin{scope}[transparency group,opacity=\opac]
			\node[line,node,light] at (y1) {};
			\node[line,node,light] at (y2) {};
			\node[line,node,light] at (y3) {};
			\node[line,node,light] at (y4) {};
			\node[line,node,light] at (y5) {};
		\end{scope}

		\path[line] (y1) to (y2);
		\path[line] (y2) to (y3);
		\path[line] (y2) to (y4);
		\path[line] (y3) to (y5);
		\path[line] (y4) to (y5);

		\pic[] at (7.5,0.6) {bar=y1-bar/1.0/0.3};
		\pic[] at (9,0.6) {bar=y2-bar/0.2/0.9};
		\pic[] at (10.5,0.6) {bar=y3-bar/0.5/0.4};
		\pic[] at (9,-2.1) {bar=y4-bar/0.5/0.4};
		\pic[] at (10.5,-2.1) {bar=y5-bar/0.7/0.8};

		\path[snake,purple] (x2) to [bend right=0] (y4);
		\path[snake,purple] (x4) to [bend right=17] (y4);
		\path[snake,purple] (x5) to (y4);

		\node[color=white,fill=white,inner sep=0pt,opacity=0.8] at (6, -0.75) {\small $f^{W_1}_{\textrm{merge}}\Big( f(v_4),f^{W_2}_{\textrm{aggr}}\big(\oms f(v_2), f(v_5) \cms \big)\!\Big)$};
		\node at (6, -0.75) {\small $f^{W_1}_{\textrm{merge}}\Big( f(v_4),f^{W_2}_{\textrm{aggr}}\big(\oms f(v_2), f(v_5) \cms \big)\!\Big)$};

	\end{tikzpicture}

	\caption{Illustration of the neighborhood aggregation step of a GNN around node $v_4$. }
	\label{gnn}
\end{figure}
Intuitively, GNNs compute a vectorial representation, i.e., a $d$-dimensional real vector, representing each node in a graph by aggregating information from neighboring nodes; see~\cref{gnn} for an illustration. Formally, let $(G,l)$ be a labeled graph with an initial node coloring $f^{(0)} \colon V(G)\rightarrow \RR^{1\times d}$ that is \emph{consistent} with $l$.
This means that each node $v$ is annotated with a feature $f^{(0)}(v)$ in $\bbR^{1\times d}$ such that $f^{(0)}(u) = f^{(0)}(v)$ if  $l(u) = l(v)$.
Alternatively, $f^{(0)}(v)$ can be an arbitrary real-valued feature vector associated with $v$, such as a cost function of a CO problem. A GNN model consists of a stack of neural network layers. Each layer aggregates local neighborhood information, i.e., neighbors' features, within each node and then passes this aggregated information to the next layer.

GNNs are often realized as follows~\citep{Morris2019a}. In each layer $t > 0$,  we compute new features
\begin{equation}\label{eq:basicgnn}
	f^{(t)}(v) = \sigma \Big( f^{(t-1)}(v) \cdot  W^{(t)}_1 +\, \sum_{\mathclap{w \in N(v)}}\,\, f^{(t-1)}(w) \cdot W_2^{(t)} \Big)
\end{equation}
in  $\bbR^{1 \times e}$ for $v$, where
$W_1^{(t)}$ and $W_2^{(t)}$ are parameter matrices from $\bbR^{d \times e}$, and $\sigma$ denotes a component-wise non-linear function, e.g., a sigmoid or a ReLU.\footnote{For clarity of presentation, we omit biases.}

Following~\cite{Gil+2017}, one may also replace the sum defined over the neighborhood in the above equation by a permutation-invariant, differentiable function. One may substitute the outer sum, e.g., by a column-wise vector concatenation.
Thus, in full generality, a new feature $f^{(t)}(v)$ is computed as
\begin{equation}\label{eq:gnngeneral}
	f^{W_1}_{\text{merge}}\Big(f^{(t-1)}(v) ,f^{W_2}_{\text{aggr}}\big(\oms f^{(t-1)}(w) \mid  w \in N(v)\cms \big)\!\Big),
\end{equation}
where $f^{W_1}_{\text{aggr}}$ aggregates over the multiset of neighborhood features and $f^{W_2}_{\text{merge}}$ merges the node's representations from step $(t-1)$ with the computed neighborhood features.
Both $f^{W_1}_{\text{aggr}}$ and $f^{W_2}_{\text{merge}}$ may be arbitrary, differentiable functions and, by analogy to \eqref{eq:basicgnn}, we denote their parameters as $W_1$ and $W_2$, respectively. To adapt the parameters $W_1$ and $W_2$ of \eqref{eq:basicgnn} and \eqref{eq:gnngeneral}, they are optimized in an end-to-end fashion (usually via stochastic gradient descent methods) together with the parameters of a neural network used for classification or regression.

\section{GNNs for Combinatorial Optimization: The State of the Art}\label{cognn}

Given that many practically relevant CO problems are \textsf{NP}-hard, it is helpful to characterize algorithms for solving them as prioritizing one of two goals. The \textit{primal} goal of finding good feasible solutions, and the \textit{dual} goal of certifying optimality or proving infeasibility. In both cases, GNNs can serve as a tool for representing problem instances, states of an iterative algorithm, or both. It is not uncommon to combine the GNN's variable or constraint representations with hand-crafted features, which would otherwise be challenging to extract automatically with the GNN. Coupled with an appropriate ML paradigm (Section~\ref{sec:mlparadigms}), GNNs have been shown to guide exact and heuristic algorithms towards finding good feasible solutions faster (Section~\ref{sec:primal}). GNNs have also been used to guide certifying optimality or infeasibility more efficiently (Section~\ref{bb}). In this case, GNNs are usually integrated with an existing complete algorithm, because an optimality certificate has in general exponential size concerning the problem description size, and it is not clear how to devise GNNs with such large outputs. Beyond using standard GNN models for CO, the emerging paradigm of \new{algorithmic reasoning} provides new perspectives on designing and training GNNs that satisfy natural invariants and properties, possibly enabling improved generalization and interpretability, as we will discuss in Section~\ref{sec:algoreasoning}.

\subsection{On the Primal Side: Finding Feasible Solutions}\label{primal}
\label{sec:primal}


We begin by discussing the use of GNNs in improving the solution-finding process in CO. 
The following practical scenarios motivate the need for quickly obtaining high-quality feasible solutions, even if without optimality or approximation guarantees.

\begin{description}
	\item[a)] \textbf{Optimality guarantees are often not needed} A practitioner may only be interested in the quality of a feasible solution in absolute terms rather than relative to the (typically unknown) optimal value of a problem instance. To assess a heuristic's suitability in this scenario, one can evaluate it on a set of instances for which the optimal value is known. However, when used on a new problem instance, the heuristic's solution can only be assessed via its (absolute) objective value. This situation arises when the CO problem of interest is practically intractable with an exact solver. For example, many vehicle routing problems admit strong MIP formulations that have an exponential number of variables or constraints, similar to the TSP formulation in Example~\ref{example:tsp}, see~\cite{toth2014vehicle}. While such problems may be solved exactly using column or constraint generation~\citep{dror1994vehicle}, a heuristic that consistently finds good solutions within a short user-defined time limit may be preferable.

	\item[b)] \textbf{Optimality is desired, but quickly finding a good solution is the priority} Because optimality is still of interest here,
	      one would like to use an exact solver that is focused on the primal side. A common use case is to take a good solution and start analyzing it manually in the current application context while the exact solver keeps running in the background. An early feasible solution allows for fast decision-making, early termination of the solver, or even revisiting the mathematical model with additional constraints that were initially ignored. MIP solvers usually provide a parameter that can be set to emphasize finding solutions quickly; see CPLEX's emphasis switch parameter for an example.\footnote{\url{https://www.ibm.com/support/knowledgecenter/SSSA5P_20.1.0/ilog.odms.cplex.help/CPLEX/Parameters/topics/MIPEmphasis.html}} Among other measures, these parameters increase the time or iterations allotted to primal heuristics at nodes of the search tree, which improves the odds of finding a good solution early on in the search.
\end{description}

Alternatively, one could also develop a custom, standalone heuristic that is executed first, providing a warm start solution to the exact solver. This simple approach is widely used and addresses both goals a) and b) simultaneously when the heuristic in question is effective for the problem of interest. This can also be done in order to obtain a high-quality first solution for initiating a local search.

Next, we will discuss various approaches that leverage GNNs in the primal setting, categorizing them according to the learning paradigms of~\cref{sec:mlparadigms}. In surveying the various works, we will touch on the following key aspects: the CO problem(s) tackled (e.g., TSP, SAT, MIP, graph coloring, etc.); the training approaches or loss functions used, with a focus on how hard constraints are satisfied; and GNN architecture choices. We do note that while in some cases the choice of architecture was intentional and well-justified, most works use one of the many interchangeable GNN architectures due to favorable empirical results or architectural novelty. This points to \textit{principled architecture design for CO} as being an impactful potential topic of future research, as discussed in~\cref{limits}.

\subsubsection{Supervised Learning}\label{subsub:lhs}
Assuming access to one or more optimal or near-optimal solutions to training instances of the CO problem of interest, a supervised learning approach is justified. 
\paragraph{One-shot solution prediction} The TSP, see~\cref{tspex}, has received substantial attention from the machine learning community following the work of~\citet{Vinyals2015}. The authors use a sequence-to-sequence ``pointer network'' (Ptr-Net) to map two-dimensional points, encoding the TSP instance, to a tour of small total length. The Ptr-Net was trained with supervised learning and thus required near-optimal solutions as labels; this may be a limiting factor when the TSP instances of interest are hard to solve and thus to label. 

\citet{prates2019learning} train a GNN in a supervised manner to predict the satisfiability of the decision version of the TSP. Small-scale instances of up to 105 cities are considered. This idea has been further extended by \citet{lemos2019graph} for the decision version of the graph coloring problem. An important limitation with this last approach is that infeasible solutions (violating some constraints) may be generated by the model. While this is expected given that these two approaches tackle decision problems, it also motivates the need to design appropriate mechanisms for handling combinatorial constraints.

\paragraph{Combining supervised models with search}
\citet{joshi2019efficient} propose the use of residual gated graph convolutional networks \citep{bresson2017residual} in a supervised manner to solve the TSP. Unlike the approach of~\citet{Vinyals2015}, the model does not output a valid TSP tour but a probability for each edge to belong to the tour. The final circuit is computed subsequently using a greedy decoding or a beam search procedure. The current limitations of GNN architectures for finding good primal solutions have been subsequently analyzed by~\citet{Joshi2020} using the TSP as a case study. It is shown that for this supervised learning approach to work well, it is preferable to combine it with the more expensive beam search. An RL policy, on the other hand, can perform as well by acting greedily. This finding is not surprising as supervised learning with a single optimal solution per training instance is inherently limited when there are multiple optima (a common situation for CO problems).

Similarly, \citet{Li+2018a} propose a supervised learning framework which, when coupled at test time with classical algorithms such as tree search and local search, was claimed to perform favorably compared to some RL methods and non-learned heuristics. \citet{Li+2018a} use Graph Convolutional Networks~\citep[GCNs]{Kip+2017}, a simple GNN architecture, on combinatorial problems that are easy to reduce to the Maximum Independent Set (MIS). A training instance is associated with a label, i.e., an optimal solution. The GCN is then trained to output multiple continuous solution predictions, and the hindsight loss function~\citep{guzman2012multiple} considers the minimum (cross-entropy) loss value across the multiple predictions. As such, the training encourages the generation of diverse solutions. At test time, these multiple (continuous) predictions are passed on to a tree search and local search in an attempt to transform them into feasible, potentially high-quality solutions. A recent criticism of~\citep{Li+2018a} by~\citet{bother2022s} reveals, through an independent and comprehensive empirical study, that the former method's reported improvements are not due to learning, and that the unsupervised approach by~\cite{ahn2020learning}, to be discussed shortly, seems to work better for the MIS. 

\paragraph{Graph matching and related problems}
Besides the TSP, \cite{Fey+2020,Yuj+2019} investigate using GNNs for graph matching. Here, graph matching refers to finding an alignment between two graphs such that a cost function is minimized, i.e., similar nodes in one graph are matched to similar nodes in the other graph. Specifically,~\citet{Yuj+2019} use a GNN architecture that learns node embeddings for each node in the two graphs and an attention score that reflects the similarity between two nodes across the two graphs. The authors propose to use pair-wise and triplet losses to train the above architecture. \citet{Fey+2020} propose a two-stage architecture for the above matching problem. In the first stage, a GNN learns a node embedding to compute a similarity score between nodes based on local neighborhoods. To fix potential misalignments due to the first stage's purely local nature, the authors propose a differentiable, iterative refinement strategy that aims to reach a consensus of matched nodes.

\citet{bai2020learning} introduce a GNN-based architecture, referred to as GraphSIM, to solve the maximum common sub-graph and the graph edit distance problems. The idea is to generate vector representations for each node in the two graphs that must be compared, then compute similarity  matrices based on the embeddings of every pair of nodes in the two graphs, and finally use a standard convolutional neural network to  obtain a similarity score between the two graphs. The training is carried out in an end-to-end fashion with supervision. 

\citet{nowak2018revised} train a GNNs in a supervised fashion to predict solutions to the Quadratic Assignment Problem (QAP). To do so, they represent QAP instances as two adjacency matrices, and use the two corresponding graphs as input to a GNN.

\paragraph{Guiding primal heuristics for MIP with supervised GNNs}
Going beyond particular CO problems and towards more general frameworks, \citet{Ding2019} explore leveraging GNNs to heuristically solve MIPs by representing them as a tripartite graph consisting of variables, constraints, and a single objective node.  Here, a variable and constraint node share an edge if the variable participates in the constraints with a non-zero coefficient. The objective shares an edge with every other node. The GNN aims to predict if a binary variable should be assigned 0 or 1. They utilize the output, i.e., a variable assignment for binary variables, of the GNN to generate either local branching global cuts~\citep{Fis+2003} or using these cuts to branch at the root node. Since the generation of labeled training data is costly, they resort to predicting so-called \emph{stable variables}, i.e., a variable whose assignment does not change over a given set of feasible solutions. Still within the local branching framework, \cite{Liu_Fischetti_Lodi_2022} use a GNN to predict the initial size of the neighborhood to be explored by the algorithm, and they leverage reinforcement learning to train a policy that dynamically adapts the neighborhood size at the subsequent local branching iterations. 

\citet{nair2020solving} propose a neighborhood search heuristic for ILPs, called neural diving, that consists of a two-step procedure. By using the bipartite graph induced by the variable constraint relationship, they first train a GNN by energy modeling to predict feasible assignments, with higher probability given to better objective values. The GNN is used to produce a tentative assignment of values, and in a second step, some of these values are thrown away, then computed again by an integer programming solver by solving the sub-ILP obtained by fixing the values of those variables that were kept. A binary classifier is trained to predict which variables should be thrown away at the second step. Notice how these supervised GNNs for MIP are all combined with some form of search that leverages the powerful MIP solver to guarantee that the hard linear constraints are satisfied.

\paragraph{SAT and related problems}
For SAT problems, \citet{Sel+2019} introduce the NeuroSAT architecture, a GNN that learns to solve SAT problems in an end-to-end fashion.
The model is directly trained to act as a satisfiability classifier, which was further investigated in~\cite{Cam+2020}, also showing that GNNs are capable of generalizing to larger random instances.
It is stressed that the NeuroSAT system is still less reliable than state-of-the-art SAT solvers.
As a suggestion for further works, the authors propose to use the NeuroSAT prediction inside a traditional SAT solver,
instead of relying on an end-to-end solving. 
This kind of integration is discussed in more detail in Section \ref{bb}.
Another improvement has been proposed by \citet{sun2020neurogift}. They conjectured
that the limitations of NeuroSAT are mainly due to the difficulty of learning an all-purpose SAT solver.
They evaluate this hypothesis by training another classifier, referred to as NeuroGIST, that is trained and tested 
only with SAT problems generated from differential cryptanalysis on the block cipher GIFT. 
Experimental results show that the new model is able to perform better than the original NeuroSAT for this
specific family of instances.

\citet{abboud2020learning} learn an algorithm for estimating the model count (i.e., the number of satisfying assignments) of a Boolean formula expressed in a disjunctive normal form. To do so, they train with supervision a GNN on graphs representing formulae and output the parameters of a Gaussian distribution. The loss is the Kullback-Leibler (KL) divergence between the predicted distribution and the ground truth distribution. 

\subsubsection{Unsupervised Learning}
The approaches in this section use GNNs to produce solutions that simultaneously and directly optimize a CO problem's objective (if one exists) and minimize a measure of constraint violation (on average across a set of training instances). As such, no optimal solutions are required for the training instances, circumventing limitations of the supervised paradigm.
\citet{Toe+2019} propose a purely unsupervised approach for solving constrained optimization problems on graphs. Thereto, they trained a GNN using an unsupervised loss function, reflecting how the current solution adheres to the constraints. This idea has been also investigated by \citet{amizadeh2018learning} for solving the circuit-SAT problem. Specifically,
instances of such a problem are Boolean circuits and can be represented as a directed acyclic graph (DAG). The authors use directly the DAG structure as an input, as opposed to typical undirected representations of SAT problems such as in NeuroSAT. The training is carried out with no supervision, and is based on an innovative loss function that, when minimized,
pushes the model to produce an assignment that yields a higher satisfiability.
\citet{xu2020tilingnn} tackle the non-periodic 2D tiling problem, which consists in covering an arbitrary 2D shape using one or more types of tiles that are given as input.
To do so, they propose a loss function containing three terms: (1) maximizing the tiling coverage of a region, (2) minimizing the tile overlap, (3) avoiding holes in the shape.
The loss is then minimized by a self-supervised approach that does not need ground-truth tiling solutions for the training. 
The problem is modeled as a graph problem and a new GNN architecture, referred to as TilinGNN, is introduced.
A large variety of 2D shapes (moons, butterflies, turtles, etc.) involving more than 2\,000 tiles can be efficiently covered by this method.

Further, \citet{Kar+2020} propose an unsupervised approach with theoretical guarantees. Concretely, they use a GNN to produce a distribution over subsets of nodes, representing a possible solution of the given graph problem\footnote{Graph Partitioning and Maximum Clique are the examples considered in \citep{Kar+2020}.}, by minimizing a probabilistic penalty loss function. To produce an integral solution, they de-randomize the continuous values, using sequential decoding, and show that this integral solution obeys the given, problem-specific constraints with high probability. This result is noteworthy as most other works referenced in this survey lack theoretical guarantees of any kind.

More recently,~\citet{duan2022augment} revisited the supervised NeuroSAT~\citep{Sel+2019} through the lens of ``contrastive learning''~\citep{chen2020simple}, a popular unsupervised learning method. First, multiple views of every (unlabelled) SAT instance was generated through \emph{label-preserving} augmentations: transformations that preserve the satisfiability of the instance. Then, the same GNN encoder of NeuroSAT was trained by maximizing the agreement between the representations of different views of the same instance \emph{(positive pairs)}, while minimizing the agreement between the representations of distinct instances \emph{(negative pairs)}. They showed that these representations can then be fine-tuned with much less labelled data, than the fully-supervised method of NeuroSAT, to produce an equally accurate predictor of satisfiability. They argued that this method can be applied to DO problems beyond SAT to help reduce the sample complexity in various downstream tasks.

\subsubsection{Reinforcement Learning for Iterative Solution Construction}
Common to the supervised and unsupervised approaches discussed thus far is that the GNN is used to produce values or scores for the decision variables being optimized over, after which a feasible solution is directly read out or a search algorithm (beam search, MIP, etc.) is seeded based on the predicted variable scores. Alternatively, iterative construction algorithms are natural for many CO problems. This makes reinforcement learning an obvious candidate for automatically deriving data-driven heuristics.

Although the use of RL in combinatorial problems had been explored much earlier, e.g., by~\citet{zhang1995reinforcement}, the work of~\citet{bello2016neural} was one of the first to combine RL with deep neural networks in the CO setting. To overcome the need for near-optimal solutions in the approach of~\citet{Vinyals2015},~\citet{bello2016neural} proposed to train Ptr-Net models using policy gradient RL methods. However, this approached failed to address a fundamental modeling limitation of Ptr-Nets: a Ptr-Net deals with sequences as its inputs and outputs, whereas a solution to the TSP has no natural ordering and is better viewed as a set of edges that form a valid tour.

\paragraph{GNNs enable RL-based policies for CO} \citet{Khalil2017} leveraged GNNs for the first time in the context of graph optimization problems, addressing this last limitation. The GNN served as the function approximator for the value function in a Deep Q-learning (DQN) formulation of CO on graphs. The authors use a Structure2Vec GNN architecture~\citep{Dai2016}, similar to \cref{eq:basicgnn}, to embed nodes of the input graph. Through the combination of GNN and DQN, a greedy node selection policy---S2V-DQN---is learned on a set of problem instances drawn from the same distribution. In this context, the TSP can be modeled as a graph problem by considering the weighted complete graph on the cities, where the edge weights are the distances between a pair of cities. A greedy node selection heuristic for the TSP iteratively selects nodes, adding the edge connecting every two consecutively selected nodes to the final tour. As such, a feasible solution is guaranteed to be obtained after $n-1$ greedy node selection steps, where the first node is chosen arbitrarily, and $n$ is the number of nodes or cities of a TSP instance. Because embedding the complete graph with a GNN can be computationally expensive and possibly unnecessary to select a suitable node, a $k$-nearest neighbor graph can be used instead of the complete graph. \citet{Khalil2017} apply the above approach to other classical graph optimization problems such as Maximum Cut (Max-Cut) and Minimum Vertex Cover (MVC).

Additionally, they extend the framework to the Set Covering Problem (SCP), in which a minimal number of sets must be selected to cover a universe of elements. While the SCP is not typically modeled as a graph problem, it can be naturally modeled as a bipartite graph, enabling the use of GNNs as in TSP, MVC, and Max-Cut. More broadly, the reducibility among \textsf{NP}-complete problems~\citep{karp1972reducibility} guarantees that a polynomial-time transformation between any two \textsf{NP}-complete problems exists. Whether such a transformation is practically tractable (e.g., a quadratic or cubic-time transformation might be considered too expensive) or whether the greedy node selection approach makes sense will depend on the particular combinatorial problem at hand. However, the approach introduced by \citet{Khalil2017} seems to be useful for a variety of problems and admits many direct extensions and improvements, some of which we will survey next.


\paragraph{The role of attention}
\citet{kool2018attention} tackle routing-type problems by training an encoder-decoder architecture using the REINFORCE RL algorithm~\citep{sutton1999policy}. This is based on Graph Attention Networks~\citep{Vel+2018}, a well-known GNN architecture. Problems tackled by~\citet{kool2018attention} include the TSP, the capacitated VRP (CVRP), the Orienteering Problem (OP), and the Prize-Collecting TSP (PCTSP). \citet{nazari2018reinforcement} also tackle the CVRP with a somewhat similar encoder-decoder approach.
\citet{deudon2018learning}, \citet{nazari2018reinforcement}, and \citet{kool2018attention} were perhaps the first to use attention-based models for routing problems. Attention-based models for routing problems allow each city to learn the importance of other cities to its own representation. This inductive bias seems to be useful for combinatorial problems. Additionally, as one moves from basic problems to richer ones, the GNN architecture's flexibility becomes more important in that it should be easy to incorporate additional characteristics of the problem. Notably, the encoder-decoder model of \cite{kool2018attention} is adjusted for each type of problem to accommodate its special characteristics, e.g., node penalties and capacities, the constraint that a feasible tour must include all nodes or the lack thereof, etc. This allows for a unified learning approach that can produce good heuristics for different optimization problems. Besides, \citet{franccois2019evaluate} have shown that the solutions obtained by \citet{deudon2018learning,kool2018attention,joshi2019efficient,Khalil2017} can be efficiently used as the first solution of a local search procedure for solving the TSP.

\paragraph{Handling hard constraints}
The problems discussed thus far in this section have constraints that are relatively easy to satisfy. For example, a feasible solution to a TSP instance is simply a tour on all nodes, implying that a constructive policy should only consider nodes, not in the current partial solution, and terminate as soon as a tour has been constructed. These requirements can be enforced by restricting the RL action space appropriately. As such, the training procedure and the GNN model need to focus exclusively on optimizing the average objective function of the combinatorial problem while enforcing these ``easy" constraints by manually constraining the action space of the RL agent. In many practical problems, some of the constraints may be trickier to satisfy. Consider the more general TSP with Time Windows~\citep[TSPTW]{savelsbergh1985local}, in which a node can only be visited within a node-specific time window. Here, edge weights should be interpreted as travel times rather than distances. It is easy to see how a constructive policy may ``get stuck" in a state or partial solution in which all actions are infeasible. \citet{ma2019combinatorial} tackle the TSPTW by augmenting the building blocks we have discussed so far (GNN with RL) with a {hierarchical} perspective. Some of the learnable parameters are responsible for generating feasible solutions, while others focus on minimizing the solution cost. Note, however, that the approach of~\cite{ma2019combinatorial} may still produce infeasible solutions, although it is reported to do so very rarely in experiments.
Also using RL, \citet{cappart2020combining} take another direction and propose to tackle problems that are hard to satisfy (such as the TSPTW) by reward shaping. The reward signal they introduce has two specific and hierarchic goals: first, finding a feasible and complete solution, and second, finding the solution minimizing the objective function among the feasible solutions.
The construction of a solution is stopped as soon as there is no action available, which corresponds to an infeasible partial solution. 
Each complete solution obtained has then the guarantee to be feasible.

\citet{ahn2020learning} propose to handle combinatorial constraints as follows.
They introduce the notion of \textit{deferred Markov Decision Process}, 
where at each iteration the agent decides whether a decision should be carried directly, or deferred to a subsequent iteration.
This enables the agent to focus on the easiest decisions, and to tackle the hardest decision only at the end, when the set of candidate solutions
is narrowed. Experiments are carried out on the maximum independent set problem and competitive results are obtained for graphs having millions of vertices.
However, the approach is limited to locally decomposable problems, 
where the feasibility constraint and the objective can be decomposed by locally connected variables. The model is trained with an actor-critic
reinforcement learning algorithm and is based on the GraphSAGE architecture \citep{Ham+2017}.

\paragraph{Other applications}

For SAT problems, \citet{yolcu2019learning} propose to encode SAT instances as an edge-labeled, bipartite graph and use a reinforcement learning approach to learn satisfying assignments inside a stochastic local search procedure, representing each clause and variable as a node. Here, a clause and a variable share an edge if the variable appears in the clause, while the edge labels indicate if a variables is negated in the corresponding clause.  They propose to use REINFORCE parameterized by a GNN on the above graph to learn a valid assignment on a variety of problems, e.g., $3$-SAT, clique detection, and graph coloring. To overcome the sparse reward nature of SAT problems, they additionally employ curriculum learning~\citep{Ben+2009}.

Within the RL framework for learning heuristics for graph problems, \cite{abe2019solving} propose to guide a Monte-Carlo Tree Search (MCTS) algorithm using a GNN, inspired by the success of AlphaGo Zero~\citep{silver2017mastering}. A similar approach appears in~\citep{drori2020learning}. \citet{Liu+2019} employ GNNs to learn chordal extensions in graphs. Specifically, they employ an on-policy imitation learning approach to imitate the minimum degree heuristic.





\begin{figure}[t]
	\centering

	\begin{tikzpicture}

		\def\bend{15}
		\def\opac{0.2}

		\tikzset{line/.style={draw,line width=1.5pt}}
		\tikzset{arrow/.style={line,->,>=stealth}}
		\tikzset{snake/.style={arrow,line width=1.3pt,decorate,decoration={snake,amplitude=1,segment length=6,post length=7}}}
		\tikzset{box/.style={dash pattern=on 5pt off 2pt,inner sep=5pt,rounded corners=3pt}}
		\tikzset{node/.style={shape=circle,inner sep=0pt,minimum width=15pt,line width=1pt}}
		\tikzset{light/.style={shading=axis,left color=white,right color=black,shading angle=-45}}

		\tikzset{font=\small}

		\draw[thick,rounded corners] ($(-3.5, -2.6)$) rectangle ($(2.5, 0.5)$) {};

		\draw[thick,rounded corners] ($(4.5, -3.6)$) rectangle ($(10.8, 0.5)$) {};

		\node[line,node,] (x1) at (0, 0) {$$};
		\node[line,node, label =  left:{$x_7 \leq 0$}] (x2) at (-1.0, -1.0) {$$};
		\node[line,node, label =  right:$x_7 \geq 1$] (x3) at (1.0, -1.0) {$$};
		\node[line,node, label =  left:{$x_4 \leq 2$}] (x4) at (-2.0, -2.0) {$$};
		\node[line,node, , very thick, fill=red!10, opacity=.2, red, label =  right:{$x_4 \geq 3$}] (x5) at (0.0, -2.0) {$$};

		\begin{scope}[transparency group,opacity=\opac]
			\node[line,node,light] at (x1) {};
			\node[line,node,light] at (x2) {};
			\node[line,node,light] at (x3) {};
			\node[line,node,light, ] at (x4) {};
			\node[line,node,light] at (x5) {};
		\end{scope}

		\path[line] (x1) to (x2);
		\path[line] (x1) to (x3);
		\path[line] (x2) to (x4);
		\path[line] (x2) to (x5);

		\path ($(3.0, -2)$)
		edge[arrow,line width=1pt, ->,bend right]
		node [below right= 6pt and -18pt,midway,scale=1.2] {\normalsize $\mathbf{a}_t = 4$}
		($(4.2, -3)$);

		\node[line,node,] (x6) at (8, 0) {$$};
		\node[line,node, label =  left:{$x_7 \leq 0$}] (x7) at (7.0, -1.0) {$$};
		\node[line,node, label =  right:$x_7 \geq 1$,red, fill=red, opacity=.2] (x8) at (9.0, -1.0) {$$};
		\node[line,node, label =  left:{$x_4 \leq 2$}] (x9) at (6.0, -2.0) {$$};
		\node[line,node, label =  right:{$x_4 \geq 3$}] (x10) at (8.0, -2.0) {$$};

		\node[line,node, label =  left:{$x_5 \leq -2$}] (x11) at (7.0, -3.0) {$$};
		\node[line,node, label =  right:{$x_5 \geq -1$}] (x12) at (9.0, -3.0) {$$};

		\begin{scope}[transparency group,opacity=\opac]
			\node[line,node,light] at (x6) {};
			\node[line,node,light] at (x7) {};
			\node[line,node,light] at (x8) {};
			\node[line,node,light] at (x9) {};
			\node[line,node,light] at (x10) {};
			\node[line,node,light] at (x11) {};
			\node[line,node,light] at (x12) {};

		\end{scope}

		\path[line] (x6) to (x7);
		\path[line] (x6) to (x8);
		\path[line] (x7) to (x9);
		\path[line] (x7) to (x10);

		\path[line] (x10) to (x11);
		\path[line] (x10) to (x12);

	\end{tikzpicture}

	\caption{Variable selection in the branch-and-bound integer programming algorithm as a MDP.}
	\label{fig:BnB-mdp}

\end{figure}
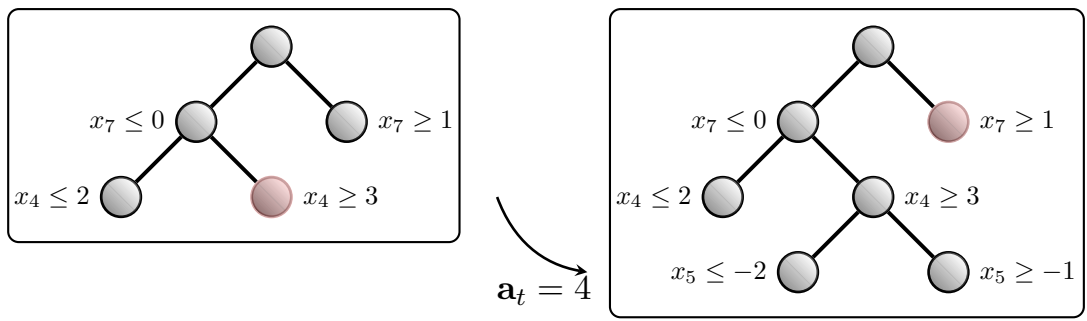

\subsection{On the Dual Side: Proving Optimality}\label{bb}

Besides finding solutions that achieve as good an objective value as possible, another common task in CO is proving that a given solution is optimal, or at least proving that the gap between the best found objective value and the optimal objective value, known as the \new{optimality gap}, is no greater than some bound. Determining such bounds is usually achieved by computing (cheap) relaxations of the optimization problem. A few works have successfully used GNNs to guide or enhance algorithms to achieve this goal. Because the task's objective is to offer proofs (of optimality or of validity of a bound), GNNs usually replace specific algorithms' components.

\subsubsection{Integer programming}
In integer linear programming, the prototypical algorithm is branch-and-bound, forming the core of all state-of-the-art solving software. Here, branching attempts to bound the optimality gap and eventually prove optimality by recursively dividing the feasible set and computing relaxations to prune away subsets that cannot contain the optimal solution. In the course of this algorithm, many decisions are repeatedly taken, whose influence is still poorly understood, and which have been described as the ``dark'' side of integer programming \citep{lodi2013heuristic}. One of the most critical is the choice, at every iteration, of the variable whose range will be divided in two. As this choice has a significant impact on the algorithm's execution time (essentially, the size of the branch-and-bound tree), there has been increased interest in learning policies, e.g., parameterized by a GNN, to select the best variable in a given context. This problem can be assimilated to the task of finding the optimal policy of a MDP, as illustrated in~\cref{fig:BnB-mdp}. 

Preliminary work had attempted the learning from fixed-dimensional representations of the problem, although without reaching the same level of performance as the best human-designed rules. In this regard, the usage of GNNs was a breakthrough, and currently represent the state of the art. The first such GNN-based approach work was the approach of \citet{Gas+2019}, who teach a GNN to imitate strong branching, an expert policy taking excellent decisions, but computationally too expensive to use in practice. The resulting policy leads to faster solving times than the solver default procedure and generalizes to larger instances than those the model is trained on.

Further work address weaknesses in this approach. One such concern was the necessity of using a GPU at inference-time. \citet{Gupta2020} propose a hybrid branching model using a GNN at the initial decision points and a light multilayer perceptron for subsequent steps, showing improvements on CPU-only hardware. Another concern is the fact that the learned policy is specialized to a type of problem, rather than being a good universal policy. \citet{nair2020solving} address the latter, first by proposing a GPU-friendly parallel linear programming solver using the alternating direction method of multipliers (ADMM) approach, and second, by showing that it could be used to scale the training to much larger datasets. In particular, they showed that they could learn a policy that performed well on MIPLIB \citep{miplib2017}, the gold standard benchmark for solvers, although at a significant computational cost. Nonetheless, it is a remarkable achievement that suggests that the GNN approach can generalize over very heterogeneous datasets, perhaps all MIPs, simply by scaling up. Finally, an alternative criticism is that the performance of the imitation learning policy will never exceed that of the expert, strong branching, which is known to perform badly on some families, such as multiple knapsack problems. This suggests that reinforcement learning methods could be a better long-term solution. The preliminary work of \citet{Sun2020} hints in this direction, by showing that evolution strategies can offer improvements over small homogeneous datasets. However, one severe limitation on using RL for branching is that,  in presence of large branch-and-bound trees, it is very difficulty to perform the so-called \emph{credit assignment}, i.e., selecting which RL actions should be credited for a specific outcome. Two recent works \citep{conf/cpaior/EtheveABJKS20, arxiv/ScavuzzoYCCGAYL22} show that the structure itself of the branch-and-bound tree can be leveraged in the attempt of simplifying the credit assignment, leading to a faster convergence of RL algorithms.

Most of the work in this area so far has focused on variable selection. Nonetheless, many other components of integer programming solvers seem just as amenable to machine learning approaches, and other aspects have started to be approached with GNN methods, such as cutting plane selection. Cutting planes are additional constraints that simplify the search process, yet are guaranteed to not remove the optimal solution, and which are continuously generated throughout the solving process. As one does not want to include all generated cutting planes by fear of rendering the computational burden unnecessarily heavy, only the most ``useful'' cuts are to be added, something which is difficult to estimate. \citet{paulus2022learning} use GNNs over a tripartite graph representation inspired by the bipartite representation of \citet{Gas+2019}, and show improvements over standard cut selection procedures in a variety of benchmarks. Also of note is the work of \citet{khalil2022mip}, where a GNN is used to predict the most likely value of variables in optimal solutions of binary integer linear programs. This, in turn, is used in their work to guide the selection of the next node to explore, as well as warm-starting the solver by rounding the predicted values combined with solver-provided ``solution repair'' heuristics.

Naturally, these works can be extended in many directions, and many other aspects of integer programming solvers seem ripe for improvement by machine learning methods, especially using GNN models. As described, many works already outperform the state-of-the-art designed by humans,  and it seems likely that solvers will integrate, over time, many such learned rules in their systems. One could in fact imagine an online training scenario, where such components would continuously learn, and solvers would improve on the type of problem of interest to the practitioner each time it solves a problem.

Finally, a closely related problem to integer linear programming is neural network verification, where a branch-and-bound algorithm is also used as backbone solution method. In fact, properties of a neural network are often verified by solving a mixed-integer optimization problem. \citet{Lu2019} represent the neural network to be verified as a graph with attributes, and train a GNN to imitate strong branching. The approach is thus close to the one of \citet{Gas+2019}, although the graphs and the GNN architecture are specifically designed for neural network verification, showing state-of-the-art improvements over hand-designed methods. Although this area has received comparatively less attention than general integer programming, their proximity suggests that there are many opportunities for translating advances in that field to neural network verification and, maybe, vice versa, too.

\subsubsection{Logic solving}

In logic solving,  such as for Boolean Satisfiability, Satisfiability Modulo Theories, and Quantified Boolean formulae, replacing human-designed decision rules by trained GNN models has also brought improvements, although the wider variety of algorithms and competing approaches make the impact less recognizable than in integer programming so far. Nonetheless, the analogies are strong enough for advances in one field to extend to the other. In particular, the role of branch-and-bound as core algorithm is instead taken by Conflict-Driven Clause Learning (CDCL), a backtracking search algorithm that resolves conflicts with resolution steps. Analogously to branch-and-bound in integer programming, in this algorithm one must repeatedly ``branch'', which in this case involves picking both an unassigned variable and a polarity (value) to assign to this variable. 

In this context, some authors have proposed representing logical formulae as graphs and using a GNN to select the best next variable, the analog of a branching step. Namely, \citet{Lederman2018} model quantified boolean formulae as bipartite graphs and teach a GNN to branch using REINFORCE. Although the reinforcement learning algorithm used is very simple, they achieve substantial improvements in number of formulae solved within a given time limit compared to VSIDS, the standard branching heuristic. Two other works apply similar ideas to related problems. \citet{Kur+2020} model propositional Boolean formulae as bipartite graphs and train a GNN to branch with $Q$-learning. Although the problem is different, they similarly show that the learned heuristic can improve on VSIDS, namely in the number of iterations needed to solve a given problem. Moreover, \citet{Vaezipoor2020} represent propositional Boolean formulae as bipartite graphs and train a GNN to branch with evolution strategies, but on a Davis–Putnam–Logemann–Loveland solver for \#SAT, the counting analog of SAT. They show that this yields improvements in solving time compared to SharpSAT, a state-of-the-art exact method. 

It is interesting to note here that these works all rely on reinforcement learning, in contrast with integer programming, where this approach, so far, is much less successful compared to imitation learning. One possible explanation is that the boolean nature of the variables and of the formulae makes the problem combinatorially less complex; another is that good experts are less obvious for these problems. Nonetheless, integer programming experience suggests that imitation learning methods, despite their drawbacks, could have a strong impact.

Finally, it is worth mentioning an approach that resembles more closely to imitation learning. We already mentioned in Section \ref{sec:primal} the work of \citet{Sel+2019} that trains a GNN ``NeuroSAT'' architecture to predict satisfiability of SAT formulae based on bipartite variable-clause graph representations. In a follow-up work, \citet{Sel+2019a} train the same architecture to predict the probability of a variable being in an unfeasible core, based on pre-solved formulae. This is used in turn to inform variable branching decisions inside the MiniSat, Glucode and Z3 SAT solvers, through the assumption that this probability correlates well with being a good variable to branch on. Using the resulting network for branching periodically, they report solving more problems on standard benchmarks than the state-of-the-art heuristic EVSIDS. Although the approach is SAT specific, it suggests that an alternative approach to variable selection can be achieved in a more indirect way by training models to predict optimal solutions, and using those predictions to guide solving, as also partially done by \cite{khalil2022mip}.

\subsubsection{Constraint programming}

The approach on variable selection found in integer programming and logic solving can also be extended to constraint programming (CP) and decision diagrams (DD). In CP, standard algorithms, such as branch-and-bound, iterative limited discrepancy search and restart-based search, are all variants of backtracking search algorithms where one must repeatedly select variables and corresponding value assignments. Value selection, in particular, has a significant impact on the quality of the search.

 In the case of constraint satisfaction programs that can be formulated as MDPs on graph states, such as the TSP with time windows, \citet{cappart2020combining} train a GNN to learn a good policy or action-value function for the Markov decision process using reinforcement learning. The resulting model is used to drive value selection within the backtracking search algorithms of CP solvers. 
 This idea is been further extended by \citet{chalumeau2021seapearl}, who propose a new CP solver that natively handles a learning component.
 To do so, they represent a CSP as a simple and undirected tripartite graph, on which each variable, possible value, and constraint are represented by nodes. The nodes are connected by an edge if and only if a variable is involved in a constraint, or if a value is inside the 
 domain of a variable. This representation has the benefit to be generic to any combinatorial problem and has
 been reused by \citet{song2022learning} for learning variable ordering heuristics.
 However, a current challenge is the size of the generated graph that can be prohibitive, making the training more tedious.
 
\subsubsection{Decision diagrams}
 
A similar situation holds with decision diagrams that are graphs that can be used to encode the feasible space of discrete problems to obtain dual bounds in CO problems \citep{bergman2016decision}. In many problems, it is possible to identify an appropriate \new{merging operator} that yields relaxed decision diagrams, whose best solution (corresponding to the shortest path in the graph) gives a dual bound. 
This mechanism is particularly interesting as the bounds obtained are flexible, meaning that their quality depends on algorithmic choices that are made during the construction of the diagram, unlike, for example, the linear relaxation bound found in integer programming.
Unfortunately, finding the algorithmic choices yielding the best bounds is often \textsf{NP}-hard. 
For instance, it is the case for determining the best variable ordering to build the diagram.
\citet{cappart2019improving} tackle this problem by training a GNN with reinforcement learning to decide which variable to add next to an incomplete decision diagram representing the problem instance that needs to be solved. 
The resulting diagram then readily yields a bound on the optimal objective value of the problem. The GNN architecture used and the problem representation as a graph is similar as the one proposed by \citet{Khalil2017}.
This idea is leveraged by \citet{parjadis2021improving} and \citet{cappart2022improving}.
They integrate the bounds obtained through this mechanism into a full-fledged branch-and-bound algorithm. The experimental results show that the learned bounds allow to reduce the solving time for the maximum independent set problem compared to methods based on non-learned bounds, 
or on linear relaxations. 

An important consideration  when designing such an approach is the increased computational power that is required  to obtain the bounds through a deep architecture.
As the model must be called many times inside the branch-and-bound tree, 
it is important that the inference is carried out efficiently. 
This issue is further discussed in \cref{limits}. This approach also suffers from some limitations, such as the need to be able to encode the problem into
a decision diagram, making this strategy not suited for solving any kind of CO problems.

\subsection{Algorithmic Reasoning}\label{algo}
\label{sec:algoreasoning}


We now turn our attention to \emph{neural algorithmic reasoning} \citep{velivckovic2021neural}, a paradigm aiming to build neural networks that align with invariants and properties of classical algorithms \citep{cormen2009introduction}. While initially conceived as a way to probe, theoretically and empirically, the extent to which neural network architectures can solve classical reasoning tasks, the area holds potential for attacking combinatorial problems over natural, noisy inputs \citep{Deac2020}. Further, it offers a framework for building neural networks in the presence of an algorithmic prior, which demonstrated significant returns for applications in pure mathematics \citep{davies2021advancing,blundell2021towards}.

Interest in such a direction persisted over many years, with several works investigating the construction of general-purpose neural computers, e.g., the neural Turing machine \citep{Graves2014}, the differentiable neural computer \citep{Graves2016}, and its variants \citep{csordas2018improving}. While such architectures have all the hallmarks of general computation, they introduce several components at once, making them often challenging to optimize, and in practice, they are almost always outperformed by simple relational reasoners \citep{Santoro2017,Santoro2018}. More carefully constructed variants and inductive biases for learning to execute \citep{Zaremba2014} have also been constructed, focusing mainly on primitive algorithms (such as arithmetic). Prominent examples include the neural GPU \citep{Kaiser2015}, neural RAM \citep{Kurach2015}, neural programmer-interpreters \citep{Reed2015}, and neural arithmetic-logic units \citep{Trask2018,Madsen2020}.

In recent times, the work in neural algorithmic reasoning has consolidated towards using graph-structured models---that is, GNNs---at the core and moving beyond primitive algorithms towards combinatorial algorithms of super-linear complexity. The primary motivation for using GNNs comes from the framework of algorithmic alignment \citep{Xu2019a,dudzik2022graph}. Through this framework, an argument is made that GNNs are potentially capable of executing polynomial-time dynamic programming algorithms \citep{bellman1966dynamic}, a paradigm from which many polynomial-time algorithms of interest can be constructed. We will investigate this framework in more detail throughout this section. 


Studying supervised algorithmic execution tasks brings up another important issue of training neural networks. Namely, (G)NNs are traditionally powerful in the \emph{interpolation} regime, i.e., when we expect the distribution of unseen (``test'') inputs to roughly match the distribution of the inputs used to train the network. However, they tend to struggle in \emph{extrapolation}, i.e., when they are evaluated out of distribution. For example, increasing the test input size, e.g., the number of nodes in the input graph, is often sufficient to lose most of the training's predictive power. Extrapolation is a potentially important issue for tackling CO problems with (G)NNs trained end-to-end. As a critical feature of a powerful reasoning system, it should apply to {any} plausible input, not just ones within the training distribution. Therefore, unless we can accurately foreshadow the kinds of inputs our neural CO approach will be solving, it could be helpful to address the issue of out-of-distribution generalization in neural networks meaningfully. 

It is also important to clarify what we mean by extrapolation in this context. Many CO problems of interest are \textsf{NP}-hard and, therefore, likely to be out of reach of GNN computation. This is because decision problems solvable by end-to-end GNNs of tractable depth (i.e., polynomial in the number of nodes) are necessarily in \textsf{P}, by definition. Furthermore, even if a problem is in \textsf{P}, a GNN needs to have sufficient depth and width to be able to solve it \citep{Loukas2020}. Hence, when we say a GNN extrapolates on a hard CO task, we will not imply that the GNN will produce optimal solutions for arbitrarily large inputs. Instead, we will require the GNN to produce solutions that align with an appropriate polynomial-time heuristic, see~\cref{align}. Similarly, extrapolation to the entire set of plausible inputs may not even be relevant. We may, instead, only care about extrapolation on a diverse\footnote{For example, it is well known that existing powerful SAT solvers can struggle over specific instances of logical formulae such as cryptography \citep{ganesh2020unreasonable}.} set of inputs that are relevant for the real-world task being solved. Training algorithmic neural networks capable of extrapolating only over those kinds of instances would still be a valuable addition to a SAT solver portfolio, even if their general extrapolation power may be substantially limited beyond those instances.

Building neural networks that extrapolate also has a natural corollary, i.e., being able to generalize across completely different kinds of problems. For example, in the work of \citep{li2018combinatorial}, a model is trained on the 3-SAT problem (by reduction to the maximal independent set problem), and it is still able to competitively solve related tasks, like finding maximal cliques or vertex covers. A better understanding of algorithmic alignment could also help us foretell which classes of related problems could benefit most from such a transfer. An example is the work on learning local heuristics from \citet{yolcu2019learning}, where training on instances of one decision problem only transfers well to problems that are \emph{similar} to it, such as vertex covering and finding dominating sets.


\subsubsection{Algorithmic alignment}\label{align}

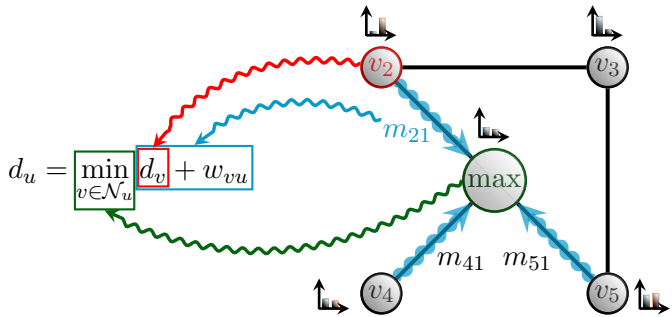
\begin{figure}
	\centering
	\begin{tikzpicture}

		\def\bend{15}
		\def\opac{0.2}

		\tikzset{line/.style={draw,line width=1.5pt}}
		\tikzset{arrow/.style={line,->,>=stealth}}
		\tikzset{snake/.style={arrow,line width=1.3pt,decorate,decoration={snake,amplitude=1,segment length=6,post length=7}}}
		\tikzset{box/.style={dash pattern=on 5pt off 2pt,inner sep=5pt,rounded corners=3pt}}
		\tikzset{node/.style={shape=circle,inner sep=0pt,minimum width=15pt,line width=1pt}}
		\tikzset{light/.style={shading=axis,left color=white,right color=black,shading angle=-45}}

		\tikzset{pics/bar/.style args={#1/#2/#3}{code={%
							\node[inner sep=0pt,minimum height=0.3cm] (#1) at (0,-0.15) {};

							\draw[fill=blue,blue] (-0.15,-0.2) rectangle (-0.075,0.25*#2-0.2);
							\draw[fill=orange,orange] (-0.025,-0.2) rectangle (0.05,0.25*#3-0.2);

							\begin{scope}[transparency group,opacity=\opac]
								\draw[fill=red,light] (-0.15,-0.2) rectangle (-0.075,0.25*#2-0.2);
								\draw[fill=orange,light] (-0.025,-0.2) rectangle (0.05,0.25*#3-0.2);
							\end{scope}

							\draw[arrow,line width=1pt] (-0.2,-0.2) -- (0.2,-0.2);
							\draw[arrow,line width=1pt] (-0.2,-0.21) -- (-0.2,0.2);
						}}}

		\node[] (bfq) {$d_u = \min\limits_{v\in\mathcal{N}_u} d_v + w_{vu}$};

		\node[line,node, right= 7em of bfq,inner sep=0.1em, green] (y3)  {$\max$};
		\node[line,node, above left=of y3, red] (y2) {$v_2$};
		\node[line,node, above right=of y3] (y1) {$v_3$};
		\node[line,node, below left=of y3] (y4) {$v_4$};
		\node[line,node, below right=of y3] (y5) {$v_5$};

		\node[rectangle, draw, thick, red, minimum height=1.3em, minimum width=1.05em] (dv) at (0.325, 0.17) {};
		\node[rectangle, draw, thick, blue, minimum height=1.5em, minimum width=4em] (msg) at (0.875, 0.17) {};
		\node[rectangle, draw, thick, green, minimum height=2em, minimum width=2.05em] (mn) at (-0.32, 0.) {};

		\begin{scope}[transparency group,opacity=\opac]
			\node[line,node,light] at (y1) {};
			\node[line,node,light] at (y2) {};
			\node[line,node,light,inner sep=0.1em] at (y3) {$\max$};
			\node[line,node,light] at (y4) {};
			\node[line,node,light] at (y5) {};
		\end{scope}

		\path[line] (y5) to (y1);
		\path[line] (y2) to node[below left,xshift=0.1em, inner sep=0.1em, blue] (Z) {$m_{21}$} (y3);
		\path[line] (y2) to (y1);
		\path[line] (y5) to node[below left,xshift=0.3em] {$m_{51}$} (y3);
		\path[line] (y4) to node[below right, xshift=-0.3em] {$m_{41}$} (y3);

		\pic[above=0.9em of y1]  {bar=y1-bar/1.0/0.3};
		\pic[above=0.9em of y2] {bar=y2-bar/0.2/0.9};
		\pic[above=0.9em of y3] {bar=y3-bar/0.5/0.4};
		\pic[left=0.9em of y4] {bar=y4-bar/0.5/0.4};
		\pic[right=0.9em of y5] {bar=y5-bar/0.7/0.8};

		\path[snake,blue,line width=3.5pt,opacity=0.7] (y2) to (y3);
		\path[snake,blue,line width=3.5pt,opacity=0.7] (y5) to (y3);
		\path[snake,blue,line width=3.5pt,opacity=0.7] (y4) to (y3);

		\path[snake,red] (y2) to [bend right] (dv.north);
		\path[snake,blue] (Z) to [bend right] (msg.north);
		\path[snake,green] (y3.west) to [bend left] (mn.south);

	\end{tikzpicture}

	\caption{Illustration of algorithmic alignment, in the case of the Bellman-Ford shortest path-finding algorithm \citep{Bellman1958}. It computes distance estimates for every node, $d_u$, and is shown on the left. Specifically, a GNN aligns well with this dynamic programming update. Node features align with intermediate computed values (\textcolor{red}{\textbf{red}}), message functions align with the candidate solutions from each neighbor (\textcolor{blue}{\textbf{blue}}), and the aggregation function (if, e.g., chosen to be $\max$) aligns with the optimization across neighbors (\textcolor{green}{\textbf{green}}).}
	\label{algo_align}
\end{figure}

The concept of \emph{algorithmic alignment} introduced by \cite{Xu2019a} is central to constructing effective algorithmic reasoners that extrapolate better. Informally, a neural network aligns with an algorithm if that algorithm can be partitioned into several {parts}, each of which can be ``easily'' modeled by one of the neural network's modules. Essentially, alignment relies on designing neural network components and control flow so that they line up well with the underlying algorithm to be learned from data. Throughout this section, we will use~\cref{algo_align} as a guiding example. Guided by this principle, novel GNN architectures and training regimes have been recently proposed to facilitate aligning with broader classes of combinatorial algorithms. As such, those works concretize the theoretical findings of \cite{Xu2019a}.

The work of \citet{Velickovic2020a} on the neural execution of graph algorithms is among the first such papers, and it focuses entirely on the practical implications of learning to execute abstract algorithmic tasks. Accordingly, it suggests several general-purpose modifications to GNNs to make them extrapolate better on combinatorial reasoning tasks.

\begin{description}
	\item[] \textbf{Using the \emph{encode-process-decode} paradigm \citep{hamrick2018relational}}  Inputs, $x$ in $\mathcal{X}$, are encoded into latents, $z$ in $\mathcal{Z}$ (e.g., $\mathcal{Z}=\mathbb{R}^d$ for $d>0$), using an encoder (G)NN, $f \colon \mathcal{X}\rightarrow\mathcal{Z}$. Latents are decoded into outputs, $y$ in $\mathcal{Y}$, using a decoder (G)NN, $g \colon \mathcal{Z}\rightarrow\mathcal{Y}$, and computation in the latent space is performed by a processor GNN, $P \colon \mathcal{Z}\rightarrow\mathcal{Z}$, which is typically executed over a certain (fixed or inferred) number of steps. Such a factorized computational model allows for easier modeling of the algorithm's \emph{intermediate state} (by decoding at different points in executing $P$). It also allows for sharing the processor across several tasks if we assume any relevant computation can be reused across them.
	\item[] \textbf{Favoring the (component-wise) max aggregation function} This aligns well with the fact  combinatorial algorithms often require some form of local decision-making, e.g., ``which neighbor is the predecessor along the shortest path?'' Moreover, max aggregation is generally more stable for larger inputs and neighborhoods. Such findings have been independently verified emperically \citep{Joshi2020,Richter2020,Corso2020} and contradict the more common advice to use  sum aggregation~\citep{Xu2019}.
	\item[] \textbf{Leveraging {strong supervision} with {teacher forcing} \citep{williams1989learning}} If, at training time, we have access to execution traces from the ground truth algorithm, which illustrates how input data is manipulated\footnote{In this sense, for sequences of unique inputs all correct sorting algorithms have the same input-output pairs, but potentially different sequences of intermediate states.} throughout that algorithm's execution, these can be used as auxiliary supervision signals, and further, the model may be asked only to predict one-step manipulations. Such an {imitation learning} setting can substantially improve out-of-distribution performance, as the additional supervision acts as a strong regularizer, constraining the function learned by the processor to more closely follow the ground-truth algorithm's iterations \citep{hussein2017imitation}. This provides a mechanism for encoding and aligning with algorithmic pre- and post-conditions, e.g., after $k$ iterations of a shortest-path algorithm such as Bellman-Ford \citep{Bellman1958}, the shortest paths that use up to $k$ hops from the source node can be computed. Strong supervision works well even without access to execution traces, as is demonstrated by the RRN model of \citet{palm2017recurrent}. Therein, the authors achieve ``convergent message passing'' by supervising a GNN to decode the ground-truth output at every step of execution.
	\item[] \textbf{Masking of outputs (and, by extension, loss functions)} GNNs are capable of processing all objects in a graph simultaneously---but for many combinatorial reasoning procedures of interest, this is unnecessary. Many efficient combinatorial algorithms are efficient precisely because they focus on only a small amount of nodes at each iteration, leaving the rest unchanged. Explicitly making the neural network predict which nodes are relevant to the current step (via a learnable {mask}, as done in \citet{Yan2020}) can therefore be impactful, and at times more important than the choice of processor.\footnote{For example, \cite{Velickovic2020a} show empirically that, for learning minimum spanning tree algorithms, LSTM processors with the masking inductive bias perform significantly better out of distribution than GNN processors without it.}
	\item[] \textbf{Executing {multiple} related algorithms} In this case, the processor network is {shared} across algorithms and becomes a multi-task learner, either simultaneously or in a curriculum \citep{Ben+2009}. When done properly, this can positively reinforce the pair-wise relations between algorithms, allowing for combining multiple heuristics into one reasoner~\citep{xhonneux2021transfer} or using the output of simpler algorithms as ``latent input'' for more complex ones, significantly improving empirical performance on the complex task.
\end{description}

While initially applied only to path-finding and spanning-tree algorithms, the prescriptions listed above have been applied for heuristically solving bipartite matching \citep{Georgiev2020}, mazes \citep{schwarzschild2021can,bansal2022end}, min-cut \citep{awasthi2021beyond}, model-based planning \citep{deac2020graph,Deac2020} and TSP \citep{Joshi2020}. It is worth noting that, concurrently, significant strides have been made on using GNNs for physics simulations \citep{Sanchez-Gonzalez2020,pfaff2020learning}. These proposals came up with a largely equivalent set of prescriptions to the ones discussed above. 

Several works have expanded on these prescriptions even further, yielding stronger classes of GNN executors. PrediNets \citep{shanahan2020explicitly} are capable of forming representations of propositions, in the context of propositional logic, and ACER \citep{georgiev2022algorithmic} further demonstrates that correct, interpretable propositional formulae can be extracted from trained PrediNets for certain tasks, such as computing minimum spanning trees. IterGNNs \citep{tang2020towards} provably align well with a broad class of algorithms that repeatedly execute some computation until a certain stopping criterion is satisfied. The design of IterGNNs allows for adaptively learning the stopping criterion, without requiring an explicit network to predict when to terminate. HomoGNNs \citep{tang2020towards} remove all biases from the GNN computation, making them align well with \emph{homogeneous functions}. These are functions exhibiting multiplicative scaling behavior---i.e. for any $\lambda$ in $\mathbb{R}$, $f(\lambda x) = \lambda f(x)$---a property held by many combinatorial tasks.\footnote{For example, if all the edge weights in a shortest path problem are multiplied by $\lambda$, \emph{any} path length---including the shortest path length---also gets multiplied by $\lambda$.} 

Recently, significant interest was also dedicated to inferring the graph over which the (G)NN should operate when solving a combinatorial task. Neural shuffle-exchange networks \citep{Freivalds2019,draguns2020residual} directly fix connectivity patterns between nodes based on results from routing theory (such as Bene\v{s} networks \citep{benevs1965mathematical}), aligning them with $O(n\log n)$ sequence processing algorithms. Lastly, pointer graph networks (PGNs) \citep{Velickovic2020} take a more pragmatic view of this issue. Rather than trying to fix a graph used by the processor GNN upfront (which may not even be given in many problems of interest), PGNs explicitly predict a graph to be used by the processor, enforcing it to match data structures' behavior.

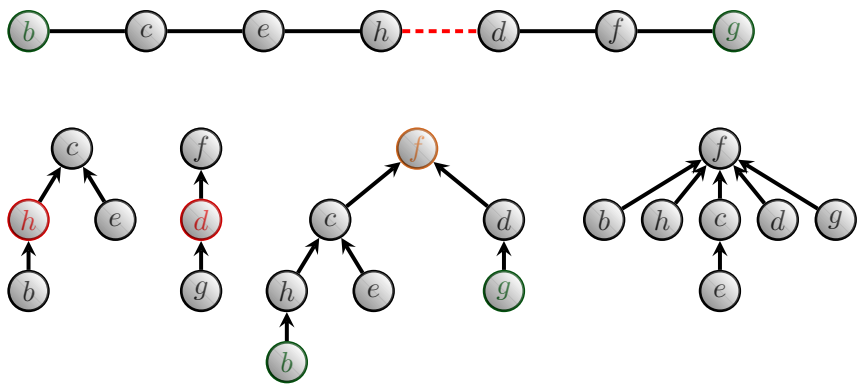
\begin{figure}
	\centering
	\begin{tikzpicture}

		\def\bend{15}
		\def\opac{0.2}

		\tikzset{line/.style={draw,line width=1.5pt}}
		\tikzset{arrow/.style={line,->,>=stealth}}
		\tikzset{snake/.style={arrow,line width=1.3pt,decorate,decoration={snake,amplitude=1,segment length=6,post length=7}}}
		\tikzset{box/.style={dash pattern=on 5pt off 2pt,inner sep=5pt,rounded corners=3pt}}
		\tikzset{node/.style={shape=circle,inner sep=0pt,minimum width=15pt,line width=1pt}}
		\tikzset{light/.style={shading=axis,left color=white,right color=black,shading angle=-45}}

		\tikzset{pics/bar/.style args={#1/#2/#3}{code={%
							\node[inner sep=0pt,minimum height=0.3cm] (#1) at (0,-0.15) {};

							\draw[fill=blue,blue] (-0.15,-0.2) rectangle (-0.075,0.25*#2-0.2);
							\draw[fill=orange,orange] (-0.025,-0.2) rectangle (0.05,0.25*#3-0.2);

							\begin{scope}[transparency group,opacity=\opac]
								\draw[fill=red,light] (-0.15,-0.2) rectangle (-0.075,0.25*#2-0.2);
								\draw[fill=orange,light] (-0.025,-0.2) rectangle (0.05,0.25*#3-0.2);
							\end{scope}

							\draw[arrow,line width=1pt] (-0.2,-0.2) -- (0.2,-0.2);
							\draw[arrow,line width=1pt] (-0.2,-0.21) -- (-0.2,0.2);
						}}}

		\node[line,node, green] (b) {$b$};
		\node[line,node, right=of b] (c) {$c$};
		\node[line,node, right=of c] (e) {$e$};
		\node[line,node, right=of e] (h) {$h$};
		\node[line,node, right=of h] (d) {$d$};
		\node[line,node, right=of d] (f) {$f$};
		\node[line,node, right=of f, green] (g) {$g$};

		\node[line, node, below=of b, xshift=1.5em] (c1) {$c$};
		\node[line, node, below=1em of c1, xshift=-1.5em, red] (h1) {$h$};
		\node[line, node, below=1em of c1, xshift=1.5em] (e1) {$e$};
		\node[line, node, below=1em of h1] (b1) {$b$};

		\node[line, node, right=3em of c1] (f1) {$f$};
		\node[line, node, below=1em of f1, red] (d1) {$d$};
		\node[line, node, below=1em of d1] (g1) {$g$};

		\node[line, node, right=6em of f1, orange] (f2) {$f$};
		\node[line, node, below=1em of f2, xshift=-3em] (c2) {$c$};
		\node[line, node, below=1em of f2, xshift=3em] (d2) {$d$};
		\node[line, node, below=1em of d2, green] (g2) {$g$};
		\node[line, node, below=1em of c2, xshift=-1.5em] (h2) {$h$};
		\node[line, node, below=1em of c2, xshift=1.5em] (e2) {$e$};
		\node[line, node, below=1em of h2, green] (b2) {$b$};

		\node[line, node, right=9em of f2] (f3) {$f$};
		\node[line, node, below=1em of f3, xshift=-4em] (b3) {$b$};
		\node[line, node, below=1em of f3, xshift=-2em] (h3) {$h$};
		\node[line, node, below=1em of f3, xshift=0em] (c3) {$c$};
		\node[line, node, below=1em of f3, xshift=2em] (d3) {$d$};
		\node[line, node, below=1em of f3, xshift=4em] (g3) {$g$};
		\node[line, node, below=1em of c3] (e3) {$e$};

		\begin{scope}[transparency group,opacity=\opac]
			\node[line,node,light] at (b) {};
			\node[line,node,light] at (c) {};
			\node[line,node,light] at (e) {};
			\node[line,node,light] at (h) {};
			\node[line,node,light] at (d) {};
			\node[line,node,light] at (f) {};
			\node[line,node,light] at (g) {};
			\node[line,node,light] at (b1) {};
			\node[line,node,light] at (c1) {};
			\node[line,node,light] at (e1) {};
			\node[line,node,light] at (h1) {};
			\node[line,node,light] at (d1) {};
			\node[line,node,light] at (f1) {};
			\node[line,node,light] at (g1) {};
			\node[line,node,light] at (b2) {};
			\node[line,node,light] at (c2) {};
			\node[line,node,light] at (e2) {};
			\node[line,node,light] at (h2) {};
			\node[line,node,light] at (d2) {};
			\node[line,node,light] at (f2) {};
			\node[line,node,light] at (g2) {};
			\node[line,node,light] at (b3) {};
			\node[line,node,light] at (c3) {};
			\node[line,node,light] at (e3) {};
			\node[line,node,light] at (h3) {};
			\node[line,node,light] at (d3) {};
			\node[line,node,light] at (f3) {};
			\node[line,node,light] at (g3) {};
		\end{scope}

		\path[line] (b) to (c);
		\path[line] (c) to (e);
		\path[line] (e) to (h);
		\path[line, red, densely dashed] (h) to (d);
		\path[line] (d) to (f);
		\path[line] (f) to (g);

		\path[line, -stealth] (h1) to (c1);
		\path[line, -stealth] (e1) to (c1);
		\path[line, -stealth] (b1) to (h1);
		\path[line, -stealth] (d1) to (f1);
		\path[line, -stealth] (g1) to (d1);

		\path[line, -stealth] (c2) to (f2);
		\path[line, -stealth] (h2) to (c2);
		\path[line, -stealth] (e2) to (c2);
		\path[line, -stealth] (b2) to (h2);
		\path[line, -stealth] (d2) to (f2);
		\path[line, -stealth] (g2) to (d2);

		\path[line, -stealth] (c3) to (f3);
		\path[line, -stealth] (h3) to (f3);
		\path[line, -stealth] (e3) to (c3);
		\path[line, -stealth] (b3) to (f3);
		\path[line, -stealth] (d3) to (f3);
		\path[line, -stealth] (g3) to (f3);

	\end{tikzpicture}

	\caption{The utility of dynamically choosing the graph to reason over for {incremental connectivity}. It is easy to construct an example \emph{path graph} (\textit{top}), wherein deciding whether one vertex is reachable from another requires linearly many GNN iterations. This can be ameliorated by reasoning over different links---namely, ones of the disjoint set union (DSU) data structure \citep{Galler1964} that represent each connected component as a \emph{rooted tree}. At the {bottom}, from {left-to-right}, we illustrate the evolution of the DSU for the graph above, once the edge $(h, d)$ is added and query $(b, g)$ executed. Note how the DSU gets {compressed} after each query \citep{Tarjan1975}, thus making it far easier for subsequent querying of whether two nodes share the same root.}
	\label{pgn_setup}
\end{figure}

As a motivating example, PGNs tackle the \emph{incremental connectivity} task (Figure \ref{pgn_setup}). Here, the model needs to answer queries of the form: given two nodes in an undirected graph, $u$ and $v$, does there exist a path between them (i.e., is $v$ reachable from $u$?)? The graph can be modified between two queries, by adding one edge at a time. It is easy to construct a worst-case ``path graph'' for which answering such queries would require $\mathrm{\Omega}(n)$ GNN steps. PGNs instead learn to imitate edges of a disjoint-set union (DSU) data structure \citep{Galler1964}. DSUs efficiently represent sets of connected components, allowing for querying reachability in $O(\alpha(n))$ amortized complexity \citep{Tarjan1975}, where $\alpha$ is the inverse Ackermann function---essentially, a {constant} for all astronomically sensible values of $n$. Thus, by carefully choosing auxiliary edges for the processor GNN, PGNs can significantly improve on the prior art in neural execution.

All of the executors listed above focus on performing message passing over exactly the nodes provided by the input graph, never modifying this node set during execution. This fundamentally limits them to simulating algorithms with up to $O(1)$ auxiliary space per node.\footnote{In reality, each node in a typical GNN stores a $d$-dimensional real vector. However, for most GNN architectures used in practice today, $d\leq 2,048$, so $O(d)$ can be treated as a constant for this analysis.} The persistent message passing (PMP) model of \citet{strathmann2021persistent} has lifted this restriction: by taking inspiration from persistent data structures \citep{driscoll1989making}, PMP allows the GNN to selectively {persist} their nodes' state after every step of message passing. Now, the nodes' latent state is never overwritten; instead, a {copy} of the persisted nodes is performed, storing their new latents. This effectively endows PMP with an episodic memory \citep{Pritzel2017} of its past computations. Further, it has the potential to overcome more general problematic aspects in learning GNNs, such as over-smoothing (see also~\cref{limits}), beyond the realm of what is possible with simple approaches like skip connections: since latent features are never overwritten, there exist no means for them to ever get smoothed out in the future.

\subsubsection{Perspectives and outlooks}


According to the previous discussion, algorithmically-aligned GNNs have been already explored in the context of dynamic programming, iterative computation,\footnote{Recent work \citep{yang2021graph} has also demonstrated that GNNs can be made to align with iterative optimization algorithms, such as proximal gradient descent and iterative reweighted least squares.} as well as algorithms backed by data structures. While novel architectures continue to be developed, there are also interesting theoretical results that further elucidate what makes an architecture extrapolate well on algorithmic execution tasks. We summarize and refer to some of these theoretical results here, and highlight two emerging outlooks on algorithmic GNNs.

Recent theoretical results have provided a unifying explanation for why algorithmically-inspired prescriptions provide benefits to extrapolating both in algorithmic and in physics-based tasks \citep{Xu2020}. Specifically, the authors make a useful geometric argument: ReLU-backed MLPs, being piece-wise linear functions, always tend to extrapolate {linearly} outside of the support of the training set. Hence, if we can design architecture components or task featurizations such that the individual parts (e.g., message functions in GNNs) have to learn roughly-linear ground-truth functions, this theoretically and practically implies stronger out-of-distribution performance. This explains, e.g., why (component-wise) max aggregation performs well for shortest path-finding. The Bellman-Ford dynamic programming rule (e.g., as in Figure \ref{algo_align})
\begin{equation}
	d_u = \min_{v\in\mathcal{N}_u} d_v + w_{vu}
\end{equation}
is an edge-wise linear function followed by a minimization. Hence, assuming a GNN of the form
\begin{equation}
	h'_u = \max_{v\in\mathcal{N}_u} M(h_u, h_v, w_{vu}),
\end{equation}
we can see that the message function $M$ now has to learn a \emph{linear function} in $h_v$ and $w_{vu}$---a substantially easier feat than if the sum-aggregation is used. Recent research has taken this insight further, demonstrating that algorithmic extrapolation may necessitate either maintaining a \emph{causal model} of the distribution shift \citep{bevilacqua2021size}, or carefully crafted self-supervised learning objectives \citep{yehudai2021local}.

While all of the above dealt with {improving} the performance of GNNs when reasoning algorithmically, for some combinatorial applications, we require the algorithmic performance to {always} remain perfect---a trait known as \emph{strong generalization} \citep{li2020strong}. Strong generalization is demonstrated to be possible. That is, {neural execution engines} (NEEs) \citep{Yan2020} are capable of empirically maintaining 100\% accuracy on various combinatorial tasks by leveraging several low-level constructs, learning individual primitive {units of computation}, such as addition, multiplication, or argmax, in isolation. Moreover, they employ {binary} representations of inputs, and conditionally mask the computation; that is, at every step, a prediction is made on which nodes are relevant, and then a GNN is executed only over the relevant nodes. Here, the focus is less on learning the algorithm itself---the dataflow between the computation units is provided in a hard-coded way, allowing for zero-shot transfer of units between related algorithms. For example, Dijkstra's algorithm \citep{dijkstra1959note} and Prim's algorithm \citep{prim1957shortest} have an identical implementation backbone---the main difference is in the key function used for a priority queue. This allows \citet{Yan2020} to directly re-use the components learnt in the context of one algorithm when learning the other.


Lastly, an important concurrent research direction that shares many insights with algorithmic reasoning is \emph{knowledge graph reasoning}. Briefly put, this area is concerned with expanding the body of knowledge in a (usually closed-domain) knowledge base, typically by inferring additional links or answering logical queries \citep{hamilton2018embedding,ren2019query2box}. Answering logical queries can often be helped by path-finding primitives, which was recently embodied in the NBFNet model \citep{zhu2021neural}. NBFNet is a knowledge graph reasoning system designed to algorithmically align to generalized versions of the previously discussed Bellman-Ford algorithm.

\subsubsection{Reasoning on natural inputs}
\label{subsub:rni}

Until now, we have focused on methodologies that allow for GNNs to strongly reason out of distribution, purely by more faithfully {imitating} existing classical algorithms. Imitation is an excellent way to {benchmark} GNN architectures for their reasoning capacity. In theory, it allows for {infinite} amounts of training or testing data of various distributions, and the fact that the underlying algorithm is known means that extrapolation can be rigorously defined.\footnote{In principle, {any} function could be a correct extrapolant if the underlying target is not known.} However, an obvious question arises: if all we are doing is imitating a classical algorithm, {why not just apply the algorithm?}

There are many potential applications of algorithmic reasoning that may provide answers to this question in principle.\footnote{Perhaps a more ``direct'' application is the ability to \emph{discover} novel algorithms. This is potentially quite promising, as most classical algorithms were constructed with a single-threaded CPU model in mind, and many of their computations may be amenable to more efficient execution on a GPU. There certainly exist preliminary signs of potential: \citet{li2020strong} detect data-driven sorting procedures that seem to improve on quicksort, and \citet{Velickovic2020} indicate, on small examples, that they are able to generalize the operations of the disjoint-set union data structure in a GPU-friendly way.} However, one particularly appealing direction for CO has already emerged---algorithmic learning executors allows us to generalize these classical combinatorial reasoners to {natural} inputs. We will thoroughly elaborate on this here.

Classical algorithms are designed with {abstraction} in mind, enforcing their inputs to conform to stringent preconditions. This is done for an apparent reason, keeping the inputs constrained enables an uninterrupted focus on ``reasoning'' and makes it far easier to certify the resulting procedure's correctness, i.e., stringent constraints. However, we must never forget why we design algorithms, to apply them to {real-world} problems. For an example of why this is at timeless odds with the way such algorithms are designed, we will look back to a 1955 study by \cite{harris1955fundamentals}, which is among the first to introduce the \emph{maximum flow} problem, before the seminal work of \cite{ford1956maximal}, and \cite{Dinic1970}, both of which present algorithms for solving it.

In line with the Cold War's contemporary issues, Harris and Ross studied the Soviet railway lines' bottleneck properties. They analyzed the rail network as a graph with edges representing railway links with scalar \emph{capacities}, corresponding to the train traffic flow rate that the railway link may support. The authors used this representation as a tool to search for the \emph{bottleneck capacity}---identifying links that would be the most effective targets for the aerial attack to disrupt the capacity maximally. Subsequent analyses have shown that this problem can be related to the \emph{minimum cut} problem on graphs and can be shown equivalent to finding a maximal flow through the network; this follows directly from the subsequently proven \emph{max-flow min-cut theorem} \citep{ford2015flows}. This problem inspired a very fruitful stream of novel combinatorial algorithms and data structures \citep{ford1956maximal,edmonds1972theoretical,Dinic1970,Sleator1983,goldberg1988new}, with applications stretching far beyond the original intent.

However, throughout their writeup, Harris and Ross remain persistently mindful of one crucial shortcoming of their proposal: the need to attach a single, scalar capacity to an entire railway link necessarily ignores a potential wealth of nuanced information from the underlying system. Quoting verbatim just one such instance:
\begin{displayquote}
	\emph{``The evaluation of both railway system and individual track capacities is, to a considerable extent, an art. The authors know of no tested mathematical model or formula that includes all of the variations and imponderables that must be weighed. Even when the individual has been closely associated with the particular territory he is evaluating, the final answer, however accurate, is largely one of judgment and experience.''}
\end{displayquote}
In many ways, this problem continues to affect applications of classical CO algorithms, being able to satisfy their preconditions necessitates converting their inputs into an abstractified form, which, if done manually, often implies drastic information loss, meaning that our combinatorial problem no longer accurately portrays the dynamics of the real world. On the other hand, the data we need to apply the algorithm may be only {partially observable}, and this can often render the algorithm completely inapplicable. 

The recent ``Amazon Last Mile Routing Research Challenge" (2021)~\citep{winkenbach2021technical} serves as early evidence of this recognition in a high-stakes setting.\footnote{\url{https://routingchallenge.mit.edu/}} The challenge is motivated by the fact that
\begin{displayquote}
\textit{``there remains an important gap between theoretical route planning and real-life route execution that most optimization-based approaches are unable to bridge. This gap relates to the fact that in real-life operations, the quality of a route is not exclusively defined by its theoretical length, duration, or cost, but by a multitude of factors that affect the extent to which drivers can effectively, safely and conveniently execute the planned route under real-life conditions."} 
\end{displayquote}
These factors involve additional contextual features and tacit knowledge (e.g., the driver's familiarity with certain routes or their observations of traffic) that would typically be dismissed if one models the routing problem using path lengths as the only data to optimize over.\footnote{It could be possible to encode this knowledge as classical CO constraints. However, the exact manner in which this can be done, especially if we want it to generalize across different drivers, is unlikely to be simple.}

On the surface, these issues appear to be fertile ground for neural networks. Their capabilities, both as a replacement for human feature engineering and a powerful processor of raw data, are highly suggestive of their potential applicability. However, this can run into several obstacles. The first one concerns learnability of such systems via gradient descent, as even if we use a neural network to encode inputs for a classical combinatorial algorithm properly, due to the discrete nature of CO problems, usual gradient-based computation is often not applicable. However, promising ways to tackle the issue of gradient estimation have already emerged\footnote{Proposals for perceptive black-box CO solvers have also emerged outside the realm of end-to-end learning; for example, \citet{brouard2020pushing} demonstrate an effective perceptive combinatorial solver by leveraging a convex formulation of graphical models.} in the literature \citep{knobelreiter2017end,wang2019satnet,vlastelica2019differentiation,mandi2020interior,niepert2021implicit}.

A more fundamental issue to consider is {data efficiency}. Even if a feasible backward pass becomes available for a combinatorial algorithm, the potential richness of raw data still needs to be bottlenecked to a scalar value. While explicitly recovering such a value allows for easier interpretability of the system, the solver is still \emph{committing} to using it; its preconditions often assume that the inputs are free of noise and estimated correctly. In contrast, neural networks derive great flexibility from their {latent representations}, that are inherently {high-dimensional},\footnote{There is a caveat that allows \emph{some} classical combinatorial algorithms to escape this bottleneck; namely, if they are \emph{designed} to operate over high-dimensional latent representations, one may just apply them out of the box to the latent representations of neural networks. A classical example is $k$-means clustering:  this insight lead \citet{wilder2019end} to propose the powerful ClusterNet model.} if any component of the neural representation ends up poorly predicted, other components are still able to step in and compensate. This is partly what enabled neural networks' emergence as a flexible tool for raw data processing. If there is insufficient data to learn how to compress it into inputs expected by the algorithm meaningfully, this may make the ultimate results of applying combinatorial algorithms on them suboptimal.

Mindful of the above, we can identify that the latest advances in neural algorithmic reasoning could lend a remarkably elegant pipeline for reasoning on natural inputs. The power comes from using the aforementioned encode-process-decode framework. Assume we have trained a GNN executor to perform a target algorithm on many abstract algorithmic inputs. The executor trained as prescribed before will have a \emph{processor network} $P$, which directly emulates one step of the algorithm, {in the latent space}.

Thus, within the weights of a properly-trained processor network, we find a polynomial-time combinatorial algorithm that is
\begin{inparaenum}
	\item[(a)] aligned with the computations of the target algorithm;
	\item[(b)] operates by matrix multiplications, hence natively admits useful gradients;
	\item[(c)] operates over high-dimensional latent spaces and hence may be more data efficient.
\end{inparaenum}

\begin{figure}
	\centering
	\scalebox{0.95}{
	\begin{tikzpicture}[scale=1.4]

		\def\bend{15}
		\def\opac{0.2}

		\tikzset{line/.style={draw,line width=1.5pt}}
		\tikzset{arrow/.style={line,->,>=stealth}}
		\tikzset{snake/.style={arrow,line width=1.3pt,decorate,decoration={snake,amplitude=1,segment length=6,post length=7}}}
		\tikzset{box/.style={dash pattern=on 5pt off 2pt,inner sep=5pt,rounded corners=3pt}}
		\tikzset{node/.style={shape=circle,inner sep=0pt,minimum width=15pt,line width=1pt}}
		\tikzset{light/.style={shading=axis,left color=white,right color=black,shading angle=-45}}

		\tikzset{pics/bar/.style args={#1/#2/#3}{code={%
							\node[inner sep=0pt,minimum height=0.3cm] (#1) at (0,-0.15) {};

							\draw[fill=blue,blue] (-0.15,-0.2) rectangle (-0.075,0.25*#2-0.2);
							\draw[fill=orange,orange] (-0.025,-0.2) rectangle (0.05,0.25*#3-0.2);

							\begin{scope}[transparency group,opacity=\opac]
								\draw[fill=red,light] (-0.15,-0.2) rectangle (-0.075,0.25*#2-0.2);
								\draw[fill=orange,light] (-0.025,-0.2) rectangle (0.05,0.25*#3-0.2);
							\end{scope}

							\draw[arrow,line width=1pt] (-0.2,-0.2) -- (0.2,-0.2);
							\draw[arrow,line width=1pt] (-0.2,-0.21) -- (-0.2,0.2);
						}}}

		\node[rectangle, draw, dashed, ultra thick, black!30, minimum height=6em, minimum width=12em] (Xbar) at (-2.8, 0.4) {};

		\node[rectangle, draw, dashed, ultra thick, black!30, minimum height=6em, minimum width=12em] (Ybar) at (5.15, 0.4) {};

		\node[rectangle, draw, dashed, ultra thick, black!30, minimum height=8.5em, minimum width=12em, fill=blue, fill opacity=0.05] (P) at (1.2, 0.4) {};

		\node[rectangle, draw, dashed, ultra thick, black!30, minimum height=6em, minimum width=12em, below=3.4em of Xbar] (X) {};

		\node[rectangle, draw, dashed, ultra thick, black!30, minimum height=6em, minimum width=12em, below=3.4em of Ybar] (Y) {};

		\node[below=0em of Xbar] (xbl) {\emph{Abstract} inputs, $\bar{x}$};

		\node[below=0em of X] (xl) {\emph{Natural} inputs, $x$};

		\node[below=0em of Ybar] (ybl) {\emph{Abstract} outputs, $\bar{y}\approx A(\bar{x})$};

		\node[below=0em of Y] (yl) {\emph{Natural} outputs, $y$};

		\draw[line width=3.5pt, -stealth, blue] (Xbar) -- node[above] {\LARGE $f$} (P);
		\draw[line width=3.5pt, -stealth, blue] (P) -- node[above] {\LARGE $g$} (Ybar);

		\draw[purple] (P) edge[loop above=1, line width=3.5pt, looseness=4.5, -stealth, opacity=0.7] (P);
		\draw[blue] (P) edge[loop above=1, line width=3.5pt, looseness=5, -stealth, opacity=0.7] node [above, opacity=1.0] {\LARGE $P$} (P);

		\draw[purple] (X) edge[bend right, line width=3.5pt, -stealth] node[below,xshift=0.1em] {\LARGE $\tilde{f}$} (P);

		\draw[purple] (P) edge[bend right, line width=3.5pt, -stealth] node[below,xshift=-0.1em] {\LARGE $\tilde{g}$} (Y);

		\foreach \pos/\name in {{(0,0)/e}, {(0,0.8)/a}, {(2.4,0)/h}, {(2.4,0.8)/d}}
		\node[line, node] (\name) at \pos {$v_{\name}$};

		\node[line, node, blue] (b) at (0.8,0.8) {$v_b$};
		\node[line, node, blue] (c) at (1.6,0.8) {$v_c$};
		\node[line, node, blue] (g) at (1.6,0) {$v_g$};
		\node[line, node, green] (f) at (0.8,0) {$v_f$};

		\foreach \pos/\name/\one/\two in {{(0.8,0.8)/b/0.8/0.3},
				{(1.6,0.8)/c/0.5/1.0}, {(0,0.8)/a/1.0/1.0}, {(2.4,0.8)/d/0.34/0.82}}
		\pic[above=0.9em of \name]  {bar=\name-bar/\one/\two};
		\foreach \pos/\name/\one/\two in {{(0.8,0)/f/0.2/0.5}, {(1.6,0)/g/0.4/0.7}, {(0,0)/e/0.9/0.4}, {(2.4,0)/h/0.6/0.5}}
		\pic[below=0.9em of \name]  {bar=\name-bar/\one/\two};

		\foreach \name/\nb in {{a/$\infty$},{c/$\infty$},{d/$\infty$},{e/$\infty$},{f/$\infty$},{g/$\infty$},{h/$\infty$}}
		\node[line, node, left = 13em of \name] (\name1) {\nb};

		\node[line, node, orange, left=13em of b] (b1) {$0$};

		\foreach \name/\nb in {{a/$\infty$},{b/$0$},{c/$\infty$},{d/$\infty$},{e/$\infty$},{f/$\infty$},{g/$\infty$},{h/$\infty$}}
		\node[line, node, below = 8em of \name1] (\name3) {$x_\name$};

		\foreach \name/\nb in {{a/1},{b/$0$},{c/2},{d/3},{e/2},{f/1},{g/2},{h/3}}
		\node[line, node, right = 13em of \name] (\name2) {\nb};

		\foreach \name/\nb in {{a/$\infty$},{b/$0$},{c/$\infty$},{d/$\infty$},{e/$\infty$},{f/$\infty$},{g/$\infty$},{h/$\infty$}}
		\node[line, node, below = 8em of \name2] (\name4) {$y_\name$};

		\begin{scope}[transparency group,opacity=\opac]
			\foreach \pos/\name in {{(0.8,0.8)/b}, {(0.8,0)/f}, {(1.6,0)/g},
					{(1.6,0.8)/c}, {(0,0)/e}, {(0,0.8)/a}, {(2.4,0)/h}, {(2.4,0.8)/d}}
			\node[line,node,light] at (\name) {};
			\foreach \pos/\name in {{(0.8,0.8)/b}, {(0.8,0)/f}, {(1.6,0)/g},
					{(1.6,0.8)/c}, {(0,0)/e}, {(0,0.8)/a}, {(2.4,0)/h}, {(2.4,0.8)/d}}
			\node[line,node,light] at (\name1) {};
			\foreach \pos/\name in {{(0.8,0.8)/b}, {(0.8,0)/f}, {(1.6,0)/g},
					{(1.6,0.8)/c}, {(0,0)/e}, {(0,0.8)/a}, {(2.4,0)/h}, {(2.4,0.8)/d}}
			\node[line,node,light] at (\name2) {};
			\foreach \pos/\name in {{(0.8,0.8)/b}, {(0.8,0)/f}, {(1.6,0)/g},
					{(1.6,0.8)/c}, {(0,0)/e}, {(0,0.8)/a}, {(2.4,0)/h}, {(2.4,0.8)/d}}
			\node[line,node,light] at (\name3) {};
			\foreach \pos/\name in {{(0.8,0.8)/b}, {(0.8,0)/f}, {(1.6,0)/g},
					{(1.6,0.8)/c}, {(0,0)/e}, {(0,0.8)/a}, {(2.4,0)/h}, {(2.4,0.8)/d}}
			\node[line,node,light] at (\name4) {};
		\end{scope}

		\path[line, black!15] (a) to (e);
		\path[line, black!15] (a) to (b);
		\path[line, black!15] (c) to (d);
		\path[line, black!15] (c) to (g);
		\path[line, black!15] (g) to (d);
		\path[line, black!15] (h) to (d);
		\path[line] (b) to (f);
		\path[line] (c) to (f);
		\path[line] (g) to (f);
		\path[line, black!15] (g) to (h);

		\path[line] (a1) to (e1);
		\path[line] (a1) to (b1);
		\path[line] (c1) to (d1);
		\path[line] (c1) to (g1);
		\path[line] (g1) to (d1);
		\path[line] (h1) to (d1);
		\path[line] (b1) to (f1);
		\path[line] (c1) to (f1);
		\path[line] (g1) to (f1);
		\path[line] (g1) to (h1);

		\path[line] (a2) to (e2);
		\path[line] (a2) to (b2);
		\path[line] (c2) to (d2);
		\path[line, black!15] (c2) to (g2);
		\path[line, black!15] (g2) to (d2);
		\path[line, black!15] (h2) to (d2);
		\path[line] (b2) to (f2);
		\path[line] (c2) to (f2);
		\path[line] (g2) to (f2);
		\path[line] (g2) to (h2);

		\path[line, red, line width=3.5pt, opacity=0.7, stealth-] (a2) to (e2);
		\path[line, red, line width=3.5pt, opacity=0.7, -stealth] (a2) to (b2);
		\path[line, red, line width=3.5pt, opacity=0.7, stealth-] (c2) to (d2);
		\path[line, red, line width=3.5pt, opacity=0.7, stealth-] (b2) to (f2);
		\path[line, red, line width=3.5pt, opacity=0.7, -stealth] (c2) to (f2);
		\path[line, red, line width=3.5pt, opacity=0.7, -stealth] (g2) to (f2);
		\path[line, red, line width=3.5pt, opacity=0.7, stealth-] (g2) to (h2);

		\path[line] (a3) to (e3);
		\path[line] (b3) to (e3);
		\path[line] (c3) to (d3);
		\path[line] (c3) to (h3);
		\path[line] (b3) to (h3);
		\path[line] (h3) to (d3);
		\path[line] (e3) to (f3);
		\path[line] (c3) to (b3);
		\path[line] (g3) to (f3);
		\path[line] (g3) to (h3);

		\path[line] (a4) to (e4);
		\path[line] (b4) to (e4);
		\path[line] (c4) to (d4);
		\path[line] (c4) to (h4);
		\path[line] (b4) to (h4);
		\path[line] (h4) to (d4);
		\path[line] (e4) to (f4);
		\path[line] (c4) to (b4);
		\path[line] (g4) to (f4);
		\path[line] (g4) to (h4);

		\path[snake,blue,line width=3.5pt,opacity=0.7] (g) to (f);
		\path[snake,blue,line width=3.5pt,opacity=0.7] (c) to (f);
		\path[snake,blue,line width=3.5pt,opacity=0.7] (b) to (f);

	\end{tikzpicture}}

	\caption{The proposed algorithmic reasoning blueprint. First, an algorithmic reasoner is trained in the encode-process-decode fashion, learning a function $g(P(f(\bar{x})))\approx  A(\bar{x})$, for a target combinatorial algorithm $A$; in this case, $A$ is breadth-first search. Once trained, the processor network $P$ is {frozen} and stitched into a pipeline over natural inputs---with new encoder and decoder $\tilde{f}$ and $\tilde{g}$. This provides an end-to-end differentiable function that has no explicit information loss, while retaining alignment with BFS.}
	\label{algo_pipe}
\end{figure}
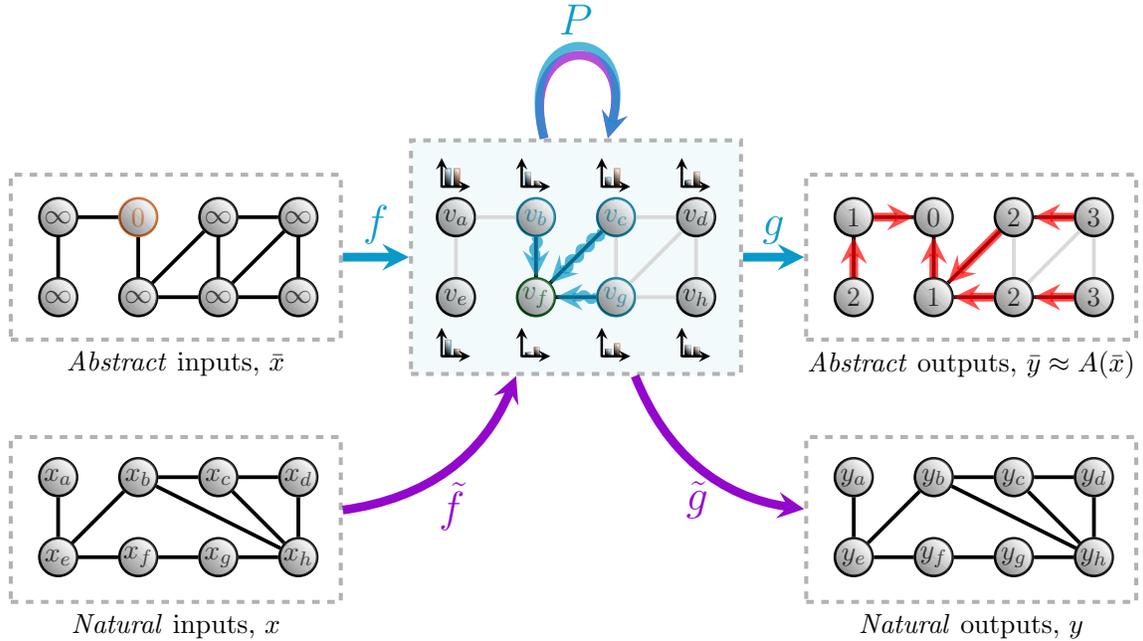

Such a processor thus seems to be a perfect component in a neural end-to-end pipeline that goes straight from raw inputs to general outputs. The general procedure for applying an algorithm $A$ (that admits abstract inputs $\bar{x}$) to raw inputs $x$ is as follows (see Figure \ref{algo_pipe}):
\begin{enumerate}
	\item Learn an algorithmic reasoner for $A$, on generated abstract inputs, $\bar{x}$, using the encode-process-decode pipeline. This yields functions $f, P, g$ such that $g(P(f(\bar{x}))) \approx A(\bar{x})$.
	\item Set up appropriate encoder and decoder neural networks, $\tilde{f}$ and $\tilde{g}$, to process raw data and produce desirable outputs.\footnote{In the case where the desired output is exactly the output of the algorithm, one may set $\tilde{g} = g$ and re-use the decoder.} The encoder should produce embeddings that correspond to the input dimension of $P$, while the decoder should operate over input embeddings that correspond to the output dimension of $P$.
	\item Swap out $f$ and $g$ for $\tilde{f}$ and $\tilde{g}$, and learn their parameters by gradient descent on any differentiable loss function that compares $\tilde{g}(P(\tilde{f}(x)))$ to ground-truth outputs, $y$. The parameters of $P$ should be kept \emph{frozen} throughout this process.
\end{enumerate}

Therefore, algorithmic reasoning presents a strong approach---through pre-trained processors\footnote{While presenting an earlier version of our work, a very important point was raised by Max Welling: if our aim is to encode a high-dimensional algorithmic solver within $P$, why not just set its weights {manually} to match the algorithm's steps? While this would certainly make $P$ trivially extrapolate, it is our belief that it would be very tricky to manually initialize it in a way that {robustly} and {diversely} uses all the dimensions of its latent input. And if $P$ only sparsely uses its latent input, we would be faced with yet another algorithmic bottleneck, limiting data efficiency. That being said, deterministic distillation of algorithms into robust high-dimensional processor networks is a potentially exciting area for future work.}---to reasoning over natural inputs. The raw encoder function $\tilde{f}$ learns how to map raw inputs onto the algorithmic input space for $P$---a task analogous to the human feature engineer---purely by backpropagation. This construction has already yielded useful architectures in the space of reinforcement learning, mainly implicit planning.

Value Iteration (VI) represents one of the most prominent model-based planning algorithms that is guaranteed to converge to an optimal RL policy. However, it requires the underlying Markov decision process to be discrete, fixed, and completely known---all requirements that are hardly satisfied in most settings of interest to deep RL. Its appeal had inspired prior work on designing neural networks that algorithmically align with VI in certain special cases, namely, in grid-worlds\footnote{Note that this does not ameliorate the requirements listed above. Assuming an environment is a grid-world places strong assumptions on the underlying MDP.} VI aligns with convolution. This yields the Value Iteration Network architecture \citep[VIN]{tamar2016value} that is a CNN-based agent, where certain convolutions share parameters, in a manner that aligns with value iteration. While this construction can outperform baseline CNN agents in terms of generalization, and its extensions to graph-based environments using GNNs have been made \citep{niu2018generalized}, the above strong constraints on the MDP remained.

In the {XLVIN} architecture, \citet{Deac2020} surpass these limitations by following precisely the algorithmic reasoning blueprint above. They pre-train an algorithmic executor for VI on several randomly-generated, known MDPs, then deploy it over a local neighborhood of the current state, derived using self-supervised learning.

The representations produced by this VI executor substantially improve a corresponding model-free RL baseline, especially in terms of data efficiency. Additionally, the model performs strongly in the low-data regime against ATreeC \citep{farquhar2017treeqn} that resorts to predicting scalar values in every node of the inferred local MDP, so that VI-style rules can be directly applied exactly according to the predictions above. As shown further by \citet{Deac2020}, even over challenging RL environments such as Atari, neurally learned algorithmic cores prove to be a viable way of applying classical combinatorial algorithms to natural inputs in a way that can surpass even a hard-coded hybrid pipeline. This is a first-view account of the potential of neural algorithmic reasoning in the real world and, given that XLVIN is only one manner in which this blueprint may see the application, we anticipate that it paves the way for many more practical CO applications.

\section{Limitations and Research Directions}\label{limits}

In the following, we give an overview of works that quantify the limitations of GNNs and the resulting implications for their use in CO. Moreover, we provide directions for further research.

\subsection{Limitations}

In the following, we survey known limitations of GNN approaches to CO.

\paragraph{Expressivity of GNNs}
Recently, different works explored the limitations of GNNs~\citep{Xu2019,Morris2019a}. Specifically,~\citet{Morris2019a} show that any GNN architecture's power to distinguish non-isomorphic graphs is upper-bounded by the $1$-dimensional Weisfeiler-Leman algorithm \citep{weisfeiler1968reduction}, a well-known polynomial-time heuristic for the graph isomorphism problem.  The heuristic is well understood and is known to have many shortcomings~\citep{Arv+2015}, such as not being able to detect cyclic information or distinguish between non-isomorphic bipartite graphs. These shortcomings have direct implications for CO applications, as they imply the existence of pairs of non-equal MIP instances that no GNN architecture can distinguish. This inspired a large body of research on stronger variants of GNNs \citep{Chen2019a,Morris2019a,Maron2019,Maron2019a,Morris2020b,Murphy2019,Murphy2019a} that provably overcome these limitations. However, such models typically do not scale to large graphs, making their usage in CO prohibitive. Alternatively, recent works~\citep{Sato2020a,Abb+2020} indicate that randomly initialized node features can help boost expressivity of GNNs, although the impact of such an approach on generalization remains unclear.

\paragraph{Generalization of GNNs}
To successfully deploy supervised machine learning models for CO, understanding generalization (out-of-training-set performance) is crucial. For example, \citet{Gar+2020} prove generalization bounds for a large class of GNNs that depend mainly on the maximum degree of the graphs, the number of layers, width, and the norms of the learned parameter matrices. Importantly, these bounds strongly depend on the sparsity of the input, which suggests that GNN's generalization ability might worsen the denser the graphs get.


\paragraph{Other limitations of GNNs}
Besides understanding generalization and expressivity in the context of combinatorial optimization, there exist other GNNs issues that might also hinder their success in the realm of CO. For example, GNNs' node features, under specific assumptions, become indistinguishable, a phenomenon referred to as \emph{over-smoothing}~\citep{Li2018}. Moreover, for GNNs, the \emph{bottleneck problem} refers to the observation that large neighborhoods cannot be accurately represented~\citep{Alon2020}. These problems prevent both methods from capturing global or long-range information, important for a number of CO problems, e.g., shortest-path applications. While there are some empirical studies~\citep{Dwivedi2020,You+2020b} investigating how different architectural design choices, e.g., skip connections or layer norm, circumvent the above mentioned problems, a theoretical understanding is still lacking.

\paragraph{Approximation and computational power}
As explained in \cref{sec:primal}, GNNs are often designed as (part of) a direct heuristic for CO tasks. Therefore, it is natural to ask what is the best {approximation ratio} achievable on various problems. By transferring results from distributed local algorithms~\citep{Suo+2013},~\citet{Sato2019} show that the best approximation ratio achievable by a large class of GNNs on the minimum vertex cover problem is 2, which is suboptimal \citep{karakostas2005better}. They also show analogous suboptimality results regarding the minimum dominated set problem and the maximum matching problem. Regarding computability,~\citet{Loukas2020} proves that some GNNs can be too small to compute some properties of graphs, such as finding their diameter or a minimum spanning tree and give minimum depth and width requirements for such tasks.

\paragraph{Large inference cost}
In some machine learning applications for CO, the inference might be repeated thousands of times. If inference is time-consuming, the overall wall-clock time of a solver or heuristic may not be reduced, even if the total number of iterations is smaller. A typical example is repeated decision making within a CO solver, e.g., branching or bounding. In this common scenario, making worse decisions fast might lead to better overall solving times than good decisions slowly. The low-degree polynomial complexity of GNN inference might be insufficient to be competitive against simpler models in this setting. \citet{Gupta2020} suggests a hybrid approach in one of these scenarios by running a full-fledged GNN once and using a suitably trained MLP to continue making decisions using the embedding computed by the GNN with additional features.

\paragraph{Data limitations in CO}
Making the common assumption that the complexity classes \textsf{NP} and \textsf{co-NP} are not equal,~\citet{Yehuda2020} show that any polynomial-time sample generator for \textsf{NP}-hard problems samples from an easier sub-problem. Under some circumstances, these sampled problems may even be trivially classifiable; for example, a classifier only checks the value of one input feature. This indicates that the observed performance metrics of current supervised approaches for intractable CO tasks may be over-inflated.  However, it remains unclear how these results translate into practice, as real-world instances of CO problems are rarely worst-case ones.

\subsection{Proposed New Directions}

To stimulate further research, we propose the following key challenges and extensions.

\paragraph{Understanding when GNNs speed up CO solvers} As outlined in the previous subsection, current GNN architectures might miss crucial structural patterns in the data, while more expressive approaches do not scale to large-scale inputs. Moreover, decisions inside CO solvers, e.g., a branching decision, are often driven by simple heuristics that are cheap to compute. Although negligible when called only a few times, resorting to a GNN inside a solver for such decisions is time consuming compared to a simple heuristic. Furthermore, internal computations inside a solver can hardly be parallelized. Hence, devising GNN architectures that scale {and} simultaneously capture essential patterns remains an open challenge. However, increased expressiveness might negatively impact generalization. Nowadays, most of the supervised approaches do not give meaningful predictive performance when evaluated on out-of-training-distribution samples. Even evaluating trained models on slightly larger graph instances often leads to a significant drop in performance. Hence, understanding the trade-offs among these three aspects remains an open challenge for deploying GNNs on CO tasks.

\paragraph{Programmatic primitives} Existing work in algorithmic reasoning has produced GNN architectures that are capable of fitting certain classes of iterative algorithms and data structures. That being said, there exist many kinds of reasoning primitives that are of high interest to CO but are still not explicitly treated by this emerging area. As only a few examples, we highlight string algorithms, very common in bioinformatics, and explicitly supporting recursive primitives for which any existing GNN executor would eventually run out of representational capacity.

\paragraph{Perceptive CO} Significant strides have already been made to use GNNs to strengthen abstractified CO pipelines. Further efforts are needed to support combinatorial reasoning over real-world inputs as most CO problems are ultimately designed as proxies for solving them. Our algorithmic reasoning section hints at a few possible blueprints for supporting this, but all of them are still in the early stages. One issue still untackled by prior research is how to meaningfully extract variables for the CO optimizer when they are not trivially given. While natural inputs pose several such challenges for the CO pipeline, it is equally important to keep in mind that ``nature is not an adversary''---even if the underlying problem is \textsf{NP}-hard, the instances provided in practice may well be effectively solvable with fast heuristics, or, in some cases, exactly. 

\paragraph{Building a generic implementation framework for GNNs for CO} Although implementation frameworks for GNNs have now emerged, it is still cumbersome to integrate GNN and machine learning into state-of-the-art solvers for the practitioners. Hence, developing a kind of modeling language for integrating ML methods that abstracts from technical details remains an open challenge and is key for adopting machine learning and GNN approaches in the real world, some of the early attempts are discussed in the next section.

\section{Implementation Frameworks}
Nowadays, there are several well-documented, open-source libraries for implementing custom GNN architectures, providing a large set of readily available models from the literature. Notable examples are PyTorch Geometric~\citep{Fey2019} and Deep Graph Library~\citep{wang2019dgl}.
Conversely, libraries to simplify the usage of machine learning in CO have also been developed.
OR-Gym~\citep{software/orgym} and OpenGraphGym~\citep{software/opengraphgym} are libraries designed to facilitate the learning of heuristics for CO problems in a similar interface to the popular OpenAI Gym library~\citep{Bro+2016}. In contrast, MIPLearn~\citep{software/miplearn} is a library that facilitates the learning of configuration parameters for CO solvers.  Ecole~\citep{Pro+2020} offers a general, extensible framework for implementing and evaluating machine learning-enhanced CO. It is also based on OpenAI Gym, and it exposes several essential decision tasks arising in general-purpose CO solvers---such as SCIP~\citep{GamrathEtal2020ZR}---as control problems over MDPs. SeaPearl~\citep{chalumeau2021seapearl} is a constraint programming solver guided by reinforcement learning and that
uses GNNs for representing training instances. Finally, research in algorithmic reasoning has recently also received a supporting benchmark dataset, namely CLRS-30 \citep{velivckovic2022clrs}. In this benchmark, graph neural networks are tasked with executing thirty diverse algorithms outlined in   \citet{cormen2009introduction}. Implementations of all relevant data generation pipelines as well as several popular models in the literature are all provided within CLRS, making it a potentially useful starting point for research in the area.

\section{Conclusions}
We gave an overview of the recent applications of GNNs for CO. To that end, we gave a concise introduction to CO, the different machine learning regimes, and GNNs. Most importantly, we surveyed primal approaches that aim at finding a heuristic or optimal solution with the help of GNNs. We then presented recent dual approaches, i.e., those that use GNNs to facilitate proving that a given solution is optimal. Moreover, we gave an overview of algorithmic reasoning, i.e., data-driven approaches aiming to overcome classical algorithms' limitations. We discussed shortcomings and research directions regarding the application of GNNs to CO. Finally, we identified a set of critical challenges to stimulate future research and advance the emerging field. We hope that our survey presents a useful handbook of graph representation learning methods, perspectives, and limitations for CO, operations research, and machine learning practitioners alike, and that its insights and principles will be helpful in spurring novel research results and future avenues.

\section*{Acknowledgements and Disclosure of Funding}
Christopher Morris is partially funded by a DFG Emmy Noether grant (468502433) and  RWTH Junior Principal Investigator Fellowship under the Excellence Strategy of the Federal Government and the Länder. We thank anonymous referees for many useful comments and interesting suggestions.

\appendix


\vskip 0.2in
\bibliography{bibliography}

\end{document}